%% file: ex_article.tex
\documentclass[final]{siamonline1116}
\usepackage{mathtools,booktabs,tabu,graphicx,mystyle,epstopdf}
\usepackage{algorithm}
\usepackage[noend]{algpseudocode}

\newcommand{\xgt}{X_0}
\newcommand{\MM}{\beta}
\newcommand{\mmx}{\alpha}
\newcommand{\pp}{\Pp^{\epsilon}}

\newcommand{\vertiii}[1]{{\left\vert\kern-0.25ex\left\vert\kern-0.25ex\left\vert #1 
		\right\vert\kern-0.25ex\right\vert\kern-0.25ex\right\vert}}

\input{ex_shared}

\ifpdf
\hypersetup{
  pdftitle={\TheTitle},
  pdfauthor={\TheAuthors}
}
\fi

\begin{document}

\maketitle

\begin{abstract}
Current popular methods for Magnetic Resonance Fingerprint (MRF) recovery are bottlenecked by the heavy computations of a matched-filtering step due to the growing size and complexity of the fingerprint dictionaries in multi-parametric quantitative MRI applications. We address this shortcoming by arranging dictionary atoms in the form of cover tree structures and adopt the corresponding fast approximate nearest neighbour searches to accelerate matched-filtering. 
For datasets belonging to smooth low-dimensional manifolds cover trees offer  
search complexities logarithmic in terms of data population. With this motivation we propose an iterative reconstruction algorithm, named CoverBLIP, to address large-size MRF problems where the fingerprint dictionary i.e. discrete manifold of Bloch responses, encodes several intrinsic NMR parameters. 
We study different forms of convergence for this algorithm and we show that provided with a notion of embedding, the inexact and non-convex iterations of CoverBLIP linearly convergence toward a near-global solution with the same order of accuracy as using exact brute-force searches. Our further examinations on both synthetic and real-world datasets and using different sampling strategies, indicates between 2 to 3 orders of magnitude reduction in total search computations. Cover trees are robust against the curse-of-dimensionality and therefore  CoverBLIP provides a notion of scalability\textemdash a consistent gain in time-accuracy performance\textemdash for searching high-dimensional atoms which may not be easily preprocessed (i.e. for dimensionality reduction) due to the increasing degrees of non-linearities appearing in the emerging multi-parametric MRF dictionaries.
\end{abstract}



\input{./sections/intro}
\input{./sections/mrf_model}
\input{./sections/coverblippart}

\input{./sections/expe}


\input{./sections/conclusion}

\appendix
\section*{Supplementary materials}
The supplementary materials include the proof of our convergence results for the inexact IPG iterations~\eqref{eq:inIP2} \textemdash including the CoverBLIP algorithm \textemdash in Section~\ref{sec:proofs_supp}. Related to our numerical experiments in Section~\ref{sec:expe_invivo} of the main article and for further clarity to readers who might not be MRI/MRF experts, 
we provide a short note in Section~\ref{sec:dcf}  on using a density compensation scheme for preconditioning MRF reconstruction problems with variable-density spiral (or other non-Cartesian sampling) readouts. 

\input{./sections/proofs}

\input{./sections/dcf}

\section*{Acknowledgments}
We thank Dr. Dan Ma (Case Western Reserve University) for providing us the real data used in~\cite{MRF}. 
We also thank Dr. Pedro G{\`o}mez and Dr. Marion Menzel for useful discussions during MG's visit to GE Healthcare Global Research in Munich.
\bibliographystyle{siamplain}
\bibliography{mybiblio}
\end{document}

%% file: ex_shared.tex
\usepackage{lipsum}
\usepackage{graphicx}
\usepackage{epstopdf}
\usepackage{mystyle,epstopdf}
\usepackage{mathtools}

\ifpdf
  \DeclareGraphicsExtensions{.eps,.pdf,.png,.jpg}
\else
  \DeclareGraphicsExtensions{.eps}
\fi

\numberwithin{theorem}{section}

\newcommand{\TheTitle}{CoverBLIP: accelerated and scalable iterative matched-filtering for Magnetic Resonance Fingerprint reconstruction} 
\newcommand{\TheAuthors}{M. Golbabaee, Z. Chen, Y. Wiaux and M. Davies}

\headers{\TheTitle}{\TheAuthors}

\title{{\TheTitle}\thanks{
		Preliminary results related to this work were appeared as a MLSP conference paper~\cite{CoverBLIP-MLSP} and an ISMRM abstract~\cite{CoverBLIPISMRM}.
\funding{This work is partly funded by the EPSRC grant
	EP/M019802/1 and the ERC C-SENSE project (ERCADG-
	2015-694888). MG is also supported by the Scottish Research
	Partnership in engineering (SPRe) Award PECRE1718/18.}}}

\author{
  Mohammad Golbabaee\thanks{Computer Science Department, The University of Bath, Bath, United Kingdom
  (\email{m.golbabaee@bath.ac.uk}).}
  \and
  Zhouye Chen\footnotemark[4] \thanks{Department of Clinical Solution, Philips Research China, Shanghai, China.}
\and
Yves Wiaux\thanks{School of Engineeering and Physical Sciences, Heriot-Watt University, Edinburgh, United Kingdom.}
  \and
 Mike Davies\thanks{School of Engineering, The University of Edinburgh, Edinburgh, United Kingdom}
}

\usepackage{amsopn}


%% file: sections/intro.tex
\section{Introduction}
\label{sec:intro}
Quantitative Magnetic Resonance Imaging (Q-MRI) provides a powerful tool for measuring various intrinsic NMR properties of tissues such as the $T1$, $T2$ and $T2^*$ relaxation times, field inhomogeneity, diffusion and perfusion~\cite{toftsmri}. 
As opposed to mainstream qualitative assessments these
\emph{absolute} physical quantities can be used for tissue or pathology
identification independent of the scanner or scanning sequences. 
Despite being the long-standing goal of the MRI community,  current quantitative approaches are extremely time-inefficient and for this reason not clinically applicable.  
Standard Q-MRI approaches (e.g.~\cite{standardT1a,standardT1b,standardT2,DESPOT12})  acquire a large sequence of images in different times and use curve-fitting tools for parameter estimation in each voxel. This procedure runs separately to estimate each parameter. The long process of acquiring
multiple fully sampled images brings serious limitation to
the conventional Q-MRI approaches to apply within a reasonable time and with an acceptable signal-to-noise ratio (SNR) and resolution.

Recently Magnetic Resonance Fingerprinting (MRF) has emerged to address this short-coming and significantly accelerate the acquisition time of Q-MRI e.g. within a couple of seconds \cite{MRF}. Three key principles are behind this new paradigm: i) applying \emph{one} excitation sequence (i.e. in one acquisition run) that simultaneously encodes many quantitative parameters of interest, ii) incorporating more complicated (and sometimes random) but shorter excitation patterns than those used in conventional Q-MRI schemes, and finally  iii) significant under-sampling of the k-space data at each temporal frame. Since the seminal work of Ma \emph{et al.}~\cite{MRF} for joint quantification of $T1,T2$ relaxation times and the off-resonance frequency $B0$, several follow-up studies have successfully extended the MRF framework to measure multitude of additional quantitative parameters such as $T2^*$, perfusion, diffusion and microvascular properties~\cite{MRFT2star,EPIT1T2star,MRF-perfusion2,MRF-perfusion,MRF-diffusion-combined,MRF-vascular1}. 
The aggressively short acquisition times used in this framework, on the other hand, introduce several algorithmic challenges at the parameter estimation stage of the \emph{MRF reconstruction problem}. 
Common approaches adopt a physical model to disambiguate the lack of sufficient spatio-temporal measurements in such a highly ill-posed inverse problem.
However, for complicated excitation patterns used in the MRF acquisitions, the (temporal) magnetic responses encoding the quantitative information are no longer following an analytic (e.g. complex exponential) model and they rather require solving Bloch differential equations~\cite{jaynes1955matrix}. The MRF framework proposes to discretize the parameter space and exhaustively simulate a large dictionary of magnetic responses (fingerprints) for all combinations of the quantized NMR parameters. This dictionary is then used
for \emph{matched-filtering} in many model-based reconstruction routines (see e.g.~\cite{MRF,FISP,SVDMRF,BLIPsiam,zhao-LR-MRF,asslander-ADMMMRF}). As occurs to any multi-parametric manifold enumeration, the main drawback of such approach is the size of this
dictionary which grows exponentially in terms of the number
of parameters and their quantization resolution. This brings a serious (scalability) limitation to the current popular schemes to be applicable in the emerging multi-parametric MRF problems, as the computational complexity of exact matched-filtering using brute-force searches grows linearly  with the dictionary size. 
%

To address this shortcoming we propose an  iterative reconstruction method with \emph{inexact} updates dubbed as
 \emph{Cover BLoch response Iterative Projection} (CoverBLIP). Our algorithm  accelerates matched-filtering steps by replacing iterative brute-force searches with fast Approximate Nearest Neighbour Searches (ANNS) based on \emph{cover tree} structures constructed off-line for a given MRF dictionary. For datasets living on smooth manifolds with low intrinsic dimension (e.g. a constant number of NMR characteristics) cover tree approximate searches are shown to have \emph{logarithmic} complexity in terms of data populations~\cite{beygelzimer2006cover}. 
 Under an  embedding assumption similar to the restricted isometry property in compressed sensing theory~\cite{DonohoCS, CRT:CS,BW:manifold,modelbasedCS,D-RIP,nam2013cosparse,Blumen,inexactipg-tit}, 
 we show that CoverBLIP iterations are able to correct the inexact updates and achieve a linear global convergence i.e. stable signal recovery. We also introduce an adaptive step-size scheme that guarantees (local) monotone convergence of CoverBLIP in general cases  e.g. when the embedding assumption does not hold. The results provided in this part apply beyond the customized MRF problem considered in this paper. We examine the reconstruction time-accuracy of the proposed method on both synthetic and \emph{in-vivo} MRF datasets with different excitation sequences and k-space sampling patterns. Our experimental results indicate superiority of CoverBLIP compared to other tested baselines. Notably, CoverBLIP achieves 2-3 orders of magnitude acceleration in conducting matched-filtering while maintaining a similar accuracy as compared to using exact iterations with brute-force searches. Unlike non-scalable fast search algorithms such as KD-trees, we show that CoverBLIP maintains this superior performance when no dimensionality-reduction preprocessing is used. This feature of robustness against the high-dimensionality of search spaces makes CoverBLIP a well-suited candidate to tackle multi-parametric MRF applications with increased non-linear dynamic complexity, where applying common subspace compression preprocessing becomes prohibitive for their unfavourable compromise in the final estimation accuracy.
 
 The rest of this paper is organized as follows: in Section~\ref{sec:related} we briefly review current popular methods for MRF reconstruction. Section~\ref{sec:MRFmodel} formulates the MRF inverse problem and discuss an iterative model-based reconstruction framework for solving this problem. In Section~\ref{sec:coverblippart} we provide an introduction to cover trees and their corresponding approximate fast search algorithm. We highlight a few complexity results useful for this work and finally we present our proposed CoverBLIP algorithm for accelerated MRF reconstruction. Section~\ref{sec:proofs} is dedicated to the proof of convergence and stability of CoverBLIP. In Section~\ref{sec:expe} we present our experimental results comparing the performance of CoverBLIP against other baselines, and finally in Section~\ref{sec:conclusions} we conclude this paper and discuss possible future directions.

\section{Related works}
\label{sec:related}
In this part we review a few algorithmic solutions proposed for solving the MRF problem. In their original paper Ma \emph{et al.}~\cite{MRF} proposed a non-iterative reconstruction scheme which consists of Fourier back-projections for all temporal slices followed by a dictionary matching step where each high-dimensional temporal voxel is compared against the atoms of a MRF dictionary. The NMR parameters corresponding to the fingerprint with highest correlation is reported for that voxel. After Fourier back-projections (of the under-sampled k-space data), one obtains highly corrupted images suffering from aliasing artefacts. The main drawback of the non-iterative Template Matching (TM) approach is to ignore this specific (aliasing) noise structure which as shown in~\cite{BLIPsiam} can lead to poor estimation accuracies for short acquisition sequences.  On the other hand researches in the area of \emph{compressed sensing} have demonstrated the efficiency of Iterative Projected Gradient (IPG) schemes for solving model-based inverse problems~\cite{ISTA,FISTA,IHTCS}. Adopted from compressed sensing literature Davies \emph{et al.}~\cite{BLIPsiam} proposed an IPG type algorithm for the MRF problem and show that iterations (which require repeated application of the TM at each iteration for projection) indeed improve parameter estimation in low data/SNR regimes.  

The large size of the MRF dictionary is however a big challenge for the runtime of matched-filtering step(s) based on brute-force searches in both approaches. This issue has been addressed in two ways. First, it has been proposed to reduce the (temporal) ambient search dimension using a few SVD bases of the MRF dictionary~\cite{SVDMRF}. The main drawback of this approach is that the MRF dictionaries contain highly non-linear structures (a low-dimensional manifold of solutions of the Bloch equations) and therefore applying a linear subspace  compression trades-off the computation time against the final accuracy of the reconstructed parameters. For instance, the Steady State
Precession (FISP) sequence~\cite{FISP} which encodes two NMR parameters (i.e. $T1$, $T2$ relaxation times) required 20 principal components for representing the corresponding search space to a reasonable accuracy, whereas an Inversion Recovery Balanced SSFP (IR-BSSFP) dictionary encoding an additional  parameter (i.e.  off-resonance frequency $B0$) cannot be accurately compressed with less than 200 components~\cite{SVDMRF}. One can imagine with the rise in applications encoding a larger number of parameters associated with the non-linear dynamics such as  $T2^*$, perfusion, diffusion and microvascular properties,  etc~\cite{MRFT2star,EPIT1T2star,MRF-perfusion2,MRF-perfusion,MRF-diffusion-combined,MRF-vascular1} this issue will get worse i.e. an exponential growth in the dictionary size without a good low-dimensional subspace representation.

A second approach incorporates hierarchical clustering for implementing fast searches over the dictionary~\cite{MRF-GRM} however it suffers from the limited accuracy of using a single step (non-iterative) matched-filtering. KD-tree searches has been proposed to accelerate matched-filtering steps within an iterative reconstruction scheme~\cite{AIRMRF}. However KD-trees are known to be non-scalable and crucially dependent on a dimensionality-reduction preprocessing step. This preprocessing might be expensive in general cases but more particular to the MRF problem, the use of the SVD subspace compression scheme (as proposed in~\cite{AIRMRF}) introduces an unfavourable compromise between the accuracy and gain in acceleration, as discussed above. Besides, the suggested inexact updates \textemdash based on a small fixed number of arithmetic operations (referred to as KD-tree \emph{checks}), but unbounded search precision, per iteration of IPG \textemdash are heuristic and not known to provide a convergence guarantee (see discussions in e.g.~\cite{inexactipg-tit} for approximation-tolerant iterations).

We propose an accelerated model-based reconstruction algorithm, CoverBLIP, with consistent performance in high-dimensional search spaces. Cover tree structures provide an important feature of robustness against the curse-of-dimensionality~\cite{Navigating,beygelzimer2006cover}. For low (intrinsic) dimensional manifold data they have provable sub-linear search complexities and in addition we show that using such approximations within an iterative scheme can still result in monotone convergence (in general) and stable global reconstruction, under an embedding assumption. While we  experimentally see that the KD-tree and cover tree searches are comparably efficient in small dimensions, 
the KD-tree performance scales considerably less effective to larger ambient dimensions compared to the achievable time-accuracy of the cover tree search.




Finally, we would like to mention a number of schemes which propose to incorporate additional low rank priors motivated by the high spatio-temporal correlations of the MRF data~\cite{zhao-LR-MRF1,zhao-LR-MRF, Eldar-LRMRF,doneva-LRMRF, asslander-ADMMMRF}. Most of these schemes are however validated on sequences encoding two (or three in~\cite{Eldar-LRMRF})  NMR parameters. As discussed above a linear subspace i.e. low rank model will not scale to multi-parametric MRF problems with increased degree of non-linearities. Moreover schemes based on singular values thresholding~\cite{zhao-LR-MRF1,Eldar-LRMRF,doneva-LRMRF}  require intensive data factorization computations at each iteration. Zhao \emph{et al.}~\cite{zhao-LR-MRF} proposed to reconstruct images in the pre-calculated SVD subspace of the MRF dictionary 
and cascade the results to a TM step for parameter estimation. CoverBLIP with a temporal compression option does the same low rank subspace reconstruction however with the benefit of faster approximate searches whose total (iterative) cost is less than a single iterated brute-force search in TM (see e.g. Figure~\ref{fig:compVSacc}). More recently a deep learning approach has been adopted for the MRF problem (see e.g. \cite{cohen-DRONE, lustig-deepMRF,deepMRF_me}), the crux of which is to approximate the match-filtering step by a compact neural network during reconstruction. The MRF dictionary is only used for training the network and not in reconstruction. 
In our numerical comparisons we exclude these approaches and focus on purely dictionary-based reconstruction baselines.

%% file: sections/mrf_model.tex
\section{MRF imaging model}
\label{sec:MRFmodel}
MRF acquisitions follow a linear spatio-temporal model:
\eql{\label{eq:forward}
	Y=P_\Omega F S (X)+\xi,
	}
where $Y\in \CC^{cm\times L}$ is the k-space measurements collected by $c$ coils at $t= 1,\hdots,L$ temporal frames and corrupted by some  noise $\xi$. The MRF image (to be recovered) is represented by a complex-valued matrix $X$ of spatio-temporal resolution $n\times L$  i.e. $n$ spatial voxels and $L$ temporal frames\footnote{The real and imaginary parts of $X$ store net magnetizations 
	across two transverse axes perpendicular to the static magnetic field.}. 
The multi-coil sensitivity operator $S:\CC^{n \times L}\rightarrow \CC^{cn \times L} $  maps each temporal frame of $X$ to $c$ weighted copies according to the sensitivity maps of $c$ head-coils used in a scanner. The sensitivity maps are identical for all temporal frames and are calculated off-line either through a separate calibration process or directly from the MRF measurements~\cite{cmap-adaptive}. Throughout whenever we consider a single coil setup $c=1$, we assume $S$ to be an identity operator (i.e. $S(X)=X$) and thus the true sensitivities are absorbed by $X$. Moreover, $F$ corresponds to a Fourier operator that maps spatial images (at each temporal frame and for each coil) to the corresponding k-space measurements. This operator might correspond to the FFT transform if a Cartesian grid is used for k-space sampling e.g. in~\cite{multishotEPI,EPIT1T2star}, or it might correspond to a Non-Uniform Fourier (NUFFT) transform~\cite{NUFFT} for non-Cartesian sampling patterns such as the variable density spirals used in~\cite{MRF, FISP}. 
Finally, $P_\Omega: \CC^{cn\times L}\rightarrow \CC^{cm\times L}$ is the sub-sampling operator with respect to a set of \emph{temporally-varying} patterns $\Omega=\bigcup_{t=1}^L\Omega_t$, where $\Omega_t$ stores $m<n$ k-space locations to be sampled at the time frame $t$. This pattern is identical for all coils at that given time frame.

The linear system~\eqref{eq:forward} is under-determined due to lack of sufficient measurements (i.e. $m<n$) which means without further  assumptions it admits infinitely many solutions 
and therefore, in order to hope for a stable MRF reconstruction one needs to incorporate efficient and restrictive priors  for this type of images. 

\subsection{Bloch dynamic model}
The main source of measurements in Q-MRI are the per-voxel net magnetization of proton dipoles obtained from dynamic rotations of the external magnetic field induced by a radio frequency (RF) coil. These excitations are in the form of a sequence of Flip Angles (FA) $\{\alpha_t\}_{t=1}^L$ applied at certain time intervals known as the \emph{repetition times} (TR) which could be a constant or varying across different time-frames $t=1,\hdots,L$. \emph{Tissues with different NMR characteristics respond distinctively to these excitations}. A qualitative MRI approach studies the contrasts between different tissues in a single time frame which is often-times dependent on the sequence type and the scanner. A Q-MRI approach rather fits a physical model to all spatio-temporal measurements and obtains the \emph{absolute} NMR characteristics of the underlying tissues, however, at the cost of significantly longer acquisition times. Standard Q-MRI approaches such as DESPOT run \emph{separate} sequences to measure one parameter at a time~\cite{standardT1a,standardT1b,standardT2,DESPOT12}. They use parameter-specific sequences that usually result in analytical time-trajectories such as 
$1-2\exp{\left(-\frac{t TR}{T1}\right)}$ or $\exp{\left(-\frac{ tTR}{T2}\right)}$ to be fitted and recover the underlying parameter (here $T1$ or $T2$) per voxel. The long process of acquiring separately 
multiple fully sampled images brings serious limitations to
standard Q-MRI approaches to apply within a reasonable
time and with an acceptable SNR and
resolution.

The MRF framework relies on a similar principle, however, it adopts more  
complicated and sometimes random excitation patterns that are able to  simultaneously encode different NMR parameters and produce more distinctive dynamic signatures  in shorter acquisition times. 
The resulting temporal trajectories no longer follow simple analytic e.g. exponential forms and they require methods for approximating the solutions of the \emph{Bloch differential equations}  which capture the overall macroscopic dynamics of per-voxel magnetizations~\cite{jaynes1955matrix}. We denote by
\eq{\Bb(\Theta; TR,TE,\alpha)\in \CC^L} 
the discrete-time Bloch response of a molecular structure with a set of intrinsic NMR parameters $\Theta$ to a specific excitation sequence of length $L$ with a given FA pattern $\alpha$, repetition TR and read-out TE times. The real and imaginary parts of $\Bb$ correspond to the amount of magnetizations across two transverse-plane components perpendicular to the external static magnetic field. For instance the IR-BSSFP sequence originally proposed for the MRF framework produces distinct magnetic responses for three parameters $\Theta=\{T1,T2,B0\}$ i.e. two relaxation times $T1$, $T2$ and the off-resonance frequency $B_0$. Recent emerging MRF applications are designing sequences encoding a larger number of NMR characteristics such as $T2^*$, diffusion, perfusion and vascular properties (see e.g.~\cite{MRFT2star,EPIT1T2star,MRF-perfusion2,MRF-perfusion,MRF-diffusion-combined,MRF-vascular1}). 


Current MRF approaches discretize through a dense sampling the parameter space $\overline \Theta := [T1]\times [T2]\times [B0]\times\hdots,$ 
 simulate off-line the Bloch equations for all parameter combinations and generate a large dictionary of fingerprints $D=\{D_j\}_{j=1}^d$ 
where, 
\eql{\label{eq:fingerprints} D_j := \Bb(\overline \Theta_j; TR,TE,\alpha)\quad \forall j=1,\hdots,d,} 
and $d=\Card(\overline \Theta)$ 
is the total number of generated fingerprints (atoms). Under the \emph{voxel purity} assumption  
each spatial voxel of the MRF image corresponds to a
specific tissue with a unique NMR parameter and would approximately match to a temporal trajectory in the fingerprint dictionary.\footnote{A number of works also consider mixture models for the MRF problem (see e.g. the supplementary part of~\cite{MRF} and a recent work~\cite{mixedMRF}), however we keep the main focus of this paper to cases where the voxel purity assumption hold.} By incorporating a notion of signal intensity in this model 
the rows of the MRF image belong to a \emph{cone} associated with the fingerprints \eqref{eq:fingerprints}. Denoting
by $X_v$ the $v$-th row of X i.e., a multi-dimensional spatial voxel, we have
\eql{\label{eq:model}
X_v \in \text{cone}(D)\quad  \forall v=1,\hdots, n,}
where the discrete cone of fingerprints is defined as follows:
\eql{\label{eq:conedef}
\text{cone}(D) := \{x \in \CC^L:\, x/\gamma \in D \quad \text{for some}\, \gamma>0\}.	
	}
Here $\gamma$ corresponds to the proton density which is generally non-uniformly distributed across spatial voxels.

\subsection{Model-based MRF reconstruction}
An important source of acceleration in the MRF acquisition process comes from the significant amount of k-space under-sampling. As a result one has to deal with solving a highly ill-posed problem to disambiguate the lack of sufficient measurements. 
The discrete Bloch model in~\eqref{eq:model} plays a critical role in regularizing the inverse problem~\eqref{eq:forward} and enabling stable MRF image reconstruction and parameter estimation. 
Following the model-based compressed sensing approaches such as~\cite{modelbasedCS,BW:manifold,MIP,inexactipg-tit}, the reconstruction problem can be cast as minimizing the measurement discrepancy \textemdash though the forward model~\eqref{eq:forward}\textemdash constrained by the per-voxel Bloch cone model:
\eql{ \label{eq:CS} \argmin_X \sum_{t=1}^L \norm{Y_t-P_{\Omega_t} FS(X_t)}_2^2 \quad s.t. \quad X_v\in \text{cone}(D) \quad \forall v=1,\hdots,n.\footnote{With a slight abuse of notation by $X_t\in \CC^n$ we refer to the MRF image at its $t$-th temporal frame i.e. the $t$-th column of $X$, whereas by $X_v$ we refer to the $v$-th row of $X$ which is an $L$-dimensional spatial voxel. Also $Y_t\in\CC^{cm}$ refers the k-space measurements collected at $t$-th repetition time.}
} 
The recovered image sequence (solution) at each spatial voxel corresponds to a fingerprint representing uniquely the underlying NMR characterizations. As appeared in compressed sensing literature~\cite{vidal-subspacecluster,meLRJS,meLRTV,intersecting-bach,TIPHSI,vidal-structruedLR}, it might be  natural to think of incorporating additional priors to promote certain spatial regularities and/or low-rank structures (i.e. accounting for the correlations between neighbouring voxels or image patches) in order to improve reconstruction, see e.g.~\cite{BLIPsiam,zhao-LR-MRF1,AIRMRF} in the MRF context. However care must be taken here, since solving a multi-constrained problem combined with the non-convex fingerprints cone~\eqref{eq:model} is often intractable and therefore despite possible empirical improvements \textemdash perhaps under good initializations\textemdash the results are likely to lack global convergence guarantees. In this paper we focus on problem~\eqref{eq:CS} constrained by the cone of fingerprints.

%

A popular approach for solving compressed sensing problems
is the Iterative Projected Gradient (IPG) algorithm~\cite{ISTA,FISTA,IHTCS}. IPG is a first-order algorithm suitable for big data applications and importantly it can also apply to globally solve problems with certain non-convex constraints~\cite{IHTCS,modelbasedCS,MIP}.
 Davies \emph{et al.}~\cite{BLIPsiam} adopted this routine for the MRF reconstruction problem and named it Bloch Response Iterative Projection (BLIP). The BLIP algorithm iterates between a gradient descent update and a (voxel-wise) model projection step:
\eql{\label{eq:blip}
	X^{k+1} = \Pp_\Cc \left(X^k - \mu_k \Aa^H\left(\Aa(X^k)-Y \right)\right), 
	}
where  $\Aa(.) :=P_{\Omega} FS(.)$ is the shorthand we use for the forward operator, $\Aa^H := S^H F^H P^H_\Omega (.)$ is the adjoint operator, $\{\mu_k\}$ is the sequence of step-sizes and $\Pp_\Cc(.)$ is the Euclidean projection operator onto the set $\Cc$ i.e.
\eql{ 
	\Pp_{\Cc}(x)\in \argmin_{x\in \Cc} \norm{x-u}_2.
	}
Note that throughout we use the shorthand $\norm{.}$ to refer to the Euclidean norm i.e. the $\ell_2$ norm of a vector or the Frobenius norm a matrix. For the MRF problem and the constraint set $\Cc$ defined by \eqref{eq:model} this projection is also called \emph{matched-filtering}. After the gradient update $Z^k := X^k - \mu_k \Aa^H\left(\Aa(X^k)-Y \right)$, the matched-filtering step $X^{k+1}=\Pp_\Cc(Z^k)$ decouples into separate cone projections for each spatial voxel $v=1,\hdots n$ and is computed as follows:
\begin{align}\label{eq:exactP}
&j^* = \argmin_j \norm{Z_v -D_j/\norm{D_j}} &\text{(nearest neighbour search)}\\
&X^{k+1}_v = \Pp_{\text{cone}(D)}(Z_v) = \gamma_v D_{j^*}& \text{(rescaling)} \label{eq:exactPs}
\end{align}
where,  $\gamma_v = \max\left( \text{real}(\langle Z_v ,D_{j^*}\rangle)/  {\norm{D_{j^*}}^2},0\right)$ is the per-voxel proton density. 

The non-iterative TM approach originally proposed in~\cite{MRF} corresponds to the first iteration of BLIP with zero initialization\footnote{Throughout we assume zero initialization $X^{k=0}=\textbf{0}$ for all iterative methods unless otherwise is specified.}. However the iterative approach has shown to be more robust against shorter excitation sequences and acquisition times, where the atoms of the fingerprint dictionary become more coherent and difficult to be distinguished~\cite{BLIPsiam}.

\subsection{Dimension-reduced subspace matched-filtering}
\label{seq:mrfdimchallenge}
The BLIP algorithm breaks down the computations involved in solving the MRF problem~\eqref{eq:CS} into two local updates namely, the gradient and projection steps for which an exact matched-filtering step e.g. by using brute-force nearest neighbour searches, has the complexity $O(nLd)$ in computation time. Discretization of the multi-parameter space often results in very large size MRF dictionaries where the number of fingerprints $d$ has an exponential relationship with the number of NMR characteristics and their quantization resolutions. 
Therefore, search strategies with linear complexity in $d$ are  a serious bottleneck to the exact matched-filtering steps at the heart of  model-based approaches for solving~\eqref{eq:CS}.

Current proposed solutions for the high dimensionality of the MRF problem rely on a (low rank) subspace compression step to reduce the matching
computations~\cite{SVDMRF,AIRMRF,asslander-ADMMMRF}. Let $V\in \CC^{L\times L}$ be the eigen-basis spanning the space of the fingerprint dictionary through the singular value decomposition (SVD) i.e. $\sum_{j=1}^d D_j(D_j)^H =V \Sigma V^H$, and $V_s\in \CC^{L\times s}$ denotes the matrix of $s$-dominant eigenvectors. By assuming high (linear) correlations between fingerprints, there exists a reasonably small number $s\ll L$ for which one would have $D_j \approx V_s \widetilde D_j$ for all $j=1,\hdots d$, where $\widetilde D_j:= V_s^H D_j\in \CC^s$ and $\widetilde D := \{\widetilde D_j\}_{j=1}^d$ are the low-dimensional proxies for the original fingerprint dictionary. With this assumption one can solve the following problem instead of~\eqref{eq:CS} in lower dimensions:
\eql{ \label{eq:CS1} \argmin_{\widetilde X\in \CC^{n\times s}} \sum_{t=1}^L \norm{Y_t-P_{\Omega_t}FS\big((\widetilde X V^H)_t\big) }_2^2 \quad s.t. \quad \tilde X_v\in \text{cone}(\widetilde D) \quad \forall v=1,\hdots,n.
} 
Note that if $D$ is low-rank and fully spanned by $V_s$ then $D=V_s\widetilde D$, $\text{cone}(D)=V_s\text{cone}(\widetilde D)$ and by a change of variable we have $X=\widetilde X V^H$, and therefore both problems~\eqref{eq:CS} and~\eqref{eq:CS1} become equivalent. 
Following the IPG routine for solving this problem, the gradient updates read
\eql{\label{eq:gradupdate}
	\widetilde Z^k= \widetilde X^k - \mu_k \Aa^H\left(\Aa(\widetilde X^kV^H)-  Y \right)V,
}
where the matched-filtering $\widetilde X_v^{k+1} = \Pp_{\text{cone}(\widetilde D)}(\widetilde Z^k_v)$ and the  corresponding searches are performed in the compressed temporal domain, directly reducing the complexity of pairwise distance calculations. 
Such a compression scheme can also reduce the gradient step computations. One can write
\begin{align}
\widetilde Z^k &:= \widetilde X^k - \mu_k \Aa^H\left(\Aa(\widetilde X^kV^H)V\right) +\mu_k \Aa^H(Y) V \nonumber\\
&= \widetilde X^k - \mu_k 
S^HF^H\Big(P^H _\Omega P_\Omega \big(FS(\widetilde X^k)V^H \big)V\Big)+\mu_k \Aa^H(Y) V. \label{eq:gradcompression}
\end{align}
The last line follows from expanding $\Aa$ and it holds since both the multi-slice Fourier transform $F$ and the coil sensitivity operator $S$ act identically across all time-frames and thus they commute with the temporal compression operators $V, V^H$. As a result, the main computations for conducting the gradient updates \eqref{eq:gradcompression} i.e., the middle term, comes from the forward-backward Fourier operations across a smaller number $s< L$ of (compressed) temporal frames plus the cost of applying compression-decompression operations $V,V^H$. 
Depending on how well a low rank model can approximate the dictionary i.e. how small would $s$ be, the overall gradient computations can drop by using such subspace compression, particularly when $F$ corresponds to expensive NUFFT transforms in non-Cartesian acquisition schemes. We empirically observe that $V,V^H$ operations would not bring a major overhead in total computations.


The idea of using subspace compressions has been applied to accelerate the brute-force searches in the single-stage TM method where the complexity of searches in compressed domain decreases to $O(nsd)$~\cite{SVDMRF}. It has been also proposed to use subspace compressions within an iterative algorithm to boost the performance of fast but non-scalable searches based on KD-trees~\cite{AIRMRF}. The applicability of this approach  is totally reliant on such a compression pre-processing since it is well understood that KD-trees are inefficient in high-dimensional (ambient) search spaces. 
Beside these advantages, we would like to remind the reader about our discussion in sections~\ref{sec:related} (see also  our numerical experiments in Section~\ref{sec:expe}), that  methodologies purely relying on subspace dimensionality-reduction are prone to 
an unfavourable compromise in their estimation accuracies when applied to multi-parametric MRF dictionaries with increased non-linear complexities and growth in data population.

%% file: sections/coverblippart.tex
\section{Accelerated MRF reconstruction with scalable tree searches}
\label{sec:coverblippart}
Accelerating the Nearest Neighbour Search (NNS) is a fundamental problem in computer science and it has a long historical literature.  
Successful proposed approaches are based on building tree structures which  hierarchically partition large datasets and then use  branch-and-bound algorithms for fast NNS (see e.g.~\cite{quadtrees,kdtrees,kmeanstrees,rtrees,balltrees,pcatrees,beygelzimer2006cover,conetrees}). KD-trees \textemdash which are the multi-dimensional generalization of binary searches \textemdash are perhaps the most widely-known  
classical structure for fast searches~\cite{kdtrees}. They consist of partitioning datasets across ambient coordinate axes and therefore do not efficiently adapt to complicated low-dimensional structures of datasets embedded into high (ambient) dimensions. A dimensionality reduction step is inevitably necessary when using KD-trees since they are non-scalable and their search complexity rapidly grows in high-dimensional problems~\cite{indyk1998breakCOD}. 
Modern search algorithms circumvent the curse-of-dimensionality by using i) tree structures that could efficiently benefit from the low \emph{intrinsic} dimensionality of natural datasets, which is a key assumption in machine learning, and ii) low-complexity algorithms for performing the search \emph{approximately} i.e. Approximate Nearest Neighbour Search (ANNS).  

In the following we briefly introduce a recent data structure known as a \emph{Cover tree}~\cite{beygelzimer2006cover} and highlight certain key properties making this structure ideal for accelerated and scalable searches within iterative MRF reconstruction. Notably for datasets with low intrinsic dimensions cover trees can achieve a logarithmic search complexity in terms of data population without needing an explicit a-priori knowledge of the data structure nor a dimensionality reduction preprocessing.

%

\subsection{Cover trees}

A cover tree is a levelled tree whose nodes are associated with points in a dataset $D=\{D_j\}_{j=1}^d$ and at different scales they form covering nets for data at multiple resolutions~\cite{beygelzimer2006cover,Navigating}. Denote by $\Ss_i \subseteq D$ the set of nodes appearing at scale $i=1,\ldots,i_{\max}$, $\Ss_0=\{D_{j_0}\}$ the tree's root and by $\sigma:=\max_{D_j \in D}\norm{D_{j_0}-D_j}$ the maximal tree coverage from the root. A cover tree structure must satisfy the following  three properties:
\begin{enumerate}
	\item Nesting: $\Ss_i \subseteq \Ss_{i+1}$, once a point $p$ appears as a node in $\Ss_i$, then every lower level in the tree has that node.
	\item Covering: every node $q\in \Ss_{i+1}$ has a parent node  $p\in \Ss_{i}$, where $\norm{p-q}\leq \sigma2^{-i}$. As a result, covering becomes finer at higher scales in a dyadic fashion. 
	
	\item Separation: nodes belonging to the same scale are separated by a minimal distance which dyadically shrinks at higher scales i.e. $\forall q,q'\in\Ss_i$ we have $\norm{q-q'}>\sigma2^{-i}$.   
\end{enumerate}  
 
Depth of the \emph{implicit} cover tree constructed with respect to the constraints above might grow very large for arbitrary datasets. Indeed we can easily verify that $i_{\max}\leq \log(\Delta(D))$, where 
\eq{
	\Delta(D):=\frac{\max \norm{p-q}}{\min \norm{p-q}}, \qquad \forall p\neq q\in D.
}
is the \emph{aspect ratio} of $D$. In practice we however only keep one copy of the nodes which do not have either parent or a child other than themselves. This \emph{explicit} representation efficiently reduces the required storage space to scale $O(d)$ linearly with data population, regardless of any (intrinsic) dimensionality assumption~\cite{beygelzimer2006cover}. 

As suggested in \cite{covertree-faster}, each node $q$ could optionally save the maximum distance to its descendants denoted by
\eq{
	\mathrm{maxdist}(q):= \max_{q'\in \text{descendant}(q)} \norm{q-q'},
} 
which provides a useful information for further acceleration of the branch-and-bound algorithm used for the search step. Note that any node $q\in \Ss_i$ appearing at scale $i$  satisfies 
	\eql{\label{eq:maxdistbound}
		\mathrm{maxdist}(q) \leq \sigma \left( 2^{-i}+2^{-i-1}	+ 2^{-i-2}+\dots \right)
	< \sigma 2^{-i+1}
}
	as a result of the covering property and therefore, one might avoid saving $\mathrm{maxdist}(.)$ values and use this upper bound instead.

{\defn{\label{def:eNN} Given a dataset $D$, a query point $p$ (which might not belong to $D$) and $\epsilon\geq 0$, then a point 
		$q\in D$ from dataset is a $(1+\epsilon)$-approximate nearest neighbour of $p$ if it holds:
		\eql{ \norm{p-q} \leq (1+\epsilon)\min_{u\in D}\norm{p-u}.}
	}}	
	Algorithm~\ref{alg:NN} details the branch-and-bound procedure for $(1+\epsilon)$-ANNS for a given cover tree structure. The proof of correctness of this algorithm is available in~\cite{beygelzimer2006cover}.
	In short, we iteratively traverse down the cover tree and at each scale we populate the set of \emph{candidates} $Q_i$ with nodes in $\Ss_i$ which could be the ancestors of the nearest neighbour solution and discard others. This refinement  uses the triangular inequality and a lower bound on the distance between the grandchildren of $Q$ to the query $p$ which is calculated based on $\mathrm{maxdist}(q), \, \forall q\in Q$. Note that the maxdist information is either previously stored during  construction of the tree or is bounded by \eqref{eq:maxdistbound}. Violating the refinement criteria at line~\ref{algline:refine} in Algorithm \ref{alg:NN} implies that $\forall q' \in \mathrm{descendant}(q)$ we would have
	\begin{align*}
		 \quad \norm{p-q'} \geq \norm{p-q} -\mathrm{maxdist}(q)
		 > \mathrm{dist}_{\min}
\end{align*}
	and therefore, $q$ cannot be an ancestor of the nearest neighbour point \textemdash because the current estimate $q_c$ would anyway provide a smaller distance to the query.  	
	At the finest scale (before stopping) we search the whole set of final candidates and report a $(1+\epsilon)$-ANNS point. Note that at each scale we only compute distances for non self-parent nodes i.e. we pass without any computation distance information of the self-parent children to finer scales. 
	
	The case $\epsilon=0$ refers to the exact tree NNS where one has to continue Algorithm~\ref{alg:NN} until the finest level of the tree.   One should distinguish between this strategy and performing a brute-force search. Although they both perform an exact NNS, the complexity of Algorithm~\ref{alg:NN} 
	is empirically shown to be way less in practical datasets. Noteworthy, although in this paper we focus on the Euclidean distance metric, cover trees are flexible to use a general notion of distance with respect to other metric spaces.
	
	\begin{algorithm}[t!]
		\begin{algorithmic}[1]
		\State \textbf{Inputs:} query point $p$, cover tree structure $\Tt$ for dataset $D$, current estimate $q_c \in D$, search inaccuracy $\epsilon\geq 0$
	\State$Q_0 = \{\Ss_0\}$ 
		\If{$q_c=\{\}$} $q_c=\Ss_0$ \EndIf		
		\State $\mathrm{dist}_{\min}=\norm{p-q_c}$
		\State{$i=0$}
\While {$i<i_{\max}$ \, \text{AND} \, $\sigma 2^{-i+1}(1+\epsilon^{-1})> \d_{\min}$\,}
			\State $Q=\left\{\text{children}(q):\, q\in Q_i \right\}$
			\State $q^* = \argmin_{q\in Q} \norm{p-q}$
			\State $\mathrm{dist} = \norm{p-q^*}$ 
			\If{$\mathrm{dist}<\mathrm{dist}_{\min}$}
			\State $\mathrm{dist}_{\min}=\mathrm{dist}$
		    \State $ q_c=q^*$ 
			\EndIf
			\State $Q_{i+1} = \left\{q\in Q: \norm{p-q}\leq \mathrm{dist}_{\min} + \mathrm{maxdist}(q) \right\}$ \label{algline:refine}
			\State $i  = i+1$			
\EndWhile	
		\Return $q_c$
	\end{algorithmic}
		\caption{\label{alg:NN}Cover tree's \textbf{$(1+\epsilon)$-ANNS}
			$\left(p, \Tt, q_c\right)$  approximate search~\cite{beygelzimer2006cover}} 
	\end{algorithm}
		
	
	\subsection{Complexity of the cover tree search}
	In the construction stage, a cover tree does not need to explicitly know the low-dimensional structure of data. However through building multi-resolution nets, several key growth properties such as the tree's explicit depth, the number of children per node, and importantly the overall search complexity are characterized by the intrinsic dimension of data~\cite{Navigating}, a notion that is referred to as the \emph{doubling dimension} introduced in~\cite{assouad,heinonen}:
	
	{\defn{\label{def:doub} Let $B(q,r)$ denotes a ball of radius $r$ centred at a point $q$ in some metric space. The doubling dimension $\dim_D(\Mm)$ of a set $\Mm$ is the smallest integer such that every ball of $\Mm$ (i.e. $\forall r>0$, $\forall q\in\Mm$, $B(q,2r)\cap \Mm$) can be covered by $2^{\dim_D(\Mm)}$  balls of half radius i.e. $B(q',r)\cap \Mm$, $q'\in \Mm$. 
		}}. 
		
The doubling dimension has several 
appealing properties, for instance we have~\cite{heinonen,Navigating}:
\begin{align}
&\dim_D(\RR^L)=\Theta(L),\\
&\dim_D(\Mm_1)\leq \dim_D(\Mm_2) \quad \text{when} \quad \Mm_1\subseteq \Mm_2,\\
&\dim(\cup_{i=1}^I \Mm_i)\leq \max_i \dim_D(\Mm_i)+\log(I).\label{eq:uniondoub}
\end{align}

Following these properties, a spare point has zero dimension and a discrete set $D$ of $d$ unstructured points has $\dim_D(D)=O(\log(d))$, independent from the ambient dimension. This dimension could be further decreased by assuming certain regular structures in practical datasets e.g.  they could belong to low-dimensional manifolds (embedded in higher ambient dimensions). It has been shown that a low-dimensional manifold $\Mm\in \RR^n$ with certain smoothness and regularity assumptions has $\dim_D(\Mm)=O(K)$ where $K\ll L$ denotes the topological dimension of the manifold~\cite{dasgupta2008}. As result a dataset $D\subseteq \Mm$ which samples a manifold with $K=O(1)$ (also 
a union of constant number of O(1)-dimensional manifolds)  would have a constant doubling dimension i.e. $\dim_D(D)=O(1)$. 

	The following theorem~\cite{Navigating,beygelzimer2006cover} bounds the complexity of cover trees approximate searches using the $(1+\epsilon)$-ANNS Algorithm~\ref{alg:NN}: 
	{\thm{\label{thm:NNcomp2} Given a query which might not belong to dataset $D$, cover tree's $(1+\epsilon)$-ANNS approximate search takes at most 
			\eql{
				2^{O(\dim_D(D))}\log \Delta(D)+(1/\epsilon)^{O(\dim_D(D))} 
			}
			computations in time.
	}}
	
	In most applications e.g. uniformly distributed datasets we have $\log (\Delta) = O(\log(d))$~\cite{Navigating}. Therefore for datasets with low dimensional structures i.e. $\dim_D=O(1)$ and by using approximations one achieves search complexities logarithmic in data population $d$, as opposed to the linear complexity of a brute-force search. 
	Table~\ref{tab:ctcomplexity} outlines the construction time, memory requirement and the search time complexities of cover trees.

		
		\begin{table}[t!]
			\centering
			\scalebox{1}{
				\begin{tabular}{lccc}
					\toprule[.2em]
					Cover tree complexity & $\dim_D=O(1)$ &  $\dim_D$ unknown\\
					\midrule[.1em]
					Construction time & $O(d\log(d))$ & $O(d^2)$\\
					Construction memory & $O(d)$ & $O(d)$ \\
					Online insertion/removal & $O(\log(d))$ & $\Omega(d)$ \\
					Approximate query time &$\log (\Delta) = O(\log(d))$
					& $\Omega(d)$\\
					\bottomrule[.2em]
				\end{tabular}
			}
			\caption{Complexity of cover trees in terms of dataset population $d$ regarding the construction time and memory, online insertion and removal operations, and approximate query time when the dataset is doubling and when no assumption on its doubling dimensionality is made (i.e worst-case complexity)~\cite{beygelzimer2006cover}.}
			\label{tab:ctcomplexity}	
		\end{table}
	
	There is a great motivation behind using cover trees searches for the MRF reconstruction problem. The manifold $\Mm:=\Bb(\Theta)\in \CC^L $ corresponding to the solutions of Bloch equations is parametrized by a small number $\Card(\Theta)\ll L$ of parameters; an observation which implies the resulting fingerprint dictionary will have a low-dimensional intrinsic structure. 
	We are currently unable to provide a complete support for claiming that the Bloch response manifolds satisfy the regularity assumptions in~\cite{dasgupta2008} and that their doubling dimensions would be a constant i.e. $\dim_D(\Bb(\Theta))=O(\Card(\Theta))=O(1)$\footnote{Examples of space-filling and non-doubling manifolds parametrized by a small number of parameters exist and are discussed in~\cite{dasgupta2008}, however we do not expect that the Bloch ODE responses result in manifolds which could fall into such pathological and highly irregular categories.},
	however, we keep this motivation to adopt cover trees searches to short-cut the heavy computations of the matched-filtering step in the MRF reconstruction. Our numerical experiments in Section~\ref{sec:expe} demonstrate that by using this structure one indeed achieves significant  accelerations and great scalability i.e. a consistent performance in high ambient dimensions unrestricted by the use of a dimensionality reduction preprocessing step.
	

	\begin{algorithm}[t!]
		\begin{algorithmic}[1]
			\State \textbf{Inputs:} k-space measurements $Y$, 
			 cover tree structure $\Tt(D)$ constructed for the normalized fingerprint dictionary $D$, forward operator $\Aa:=P_\Omega F S$ and its corresponding adjoint operator $\Aa^H$, initial step-size $\mu$.
			\State \textbf{Initialization:} $k=0,\,X^0=\mathbf{0},\, \mu_k=\mu\,\, \forall k=1,2,\hdots$
			\While {stopping criterion = false }
			\State $Z = X^k - \mu_k \Aa^H\left(\Aa(X^k)-Y \right)$ \qquad\qquad\qquad\qquad\qquad \#(gradient update)
			\For{$v=1,\hdots,n$}{\qquad\qquad\qquad\qquad\qquad \#(per-voxel approximate model projection)}
			\State $D_{j^*_{k+1},v} = \textbf{$(1+\epsilon)$-ANNS}\left( Z_v/\norm{Z_v}, \Tt(D), D_{j^*_{k},v}\right)$ \, \#(cover tree's ANNS)\label{algline:6}
			\State $\gamma_v = \max\left( \text{real}(\langle Z_v ,D_{j^*_{k+1},v}\rangle)/  {\norm{D_{j^*_{k+1},v}}^2},0\right)$ 
			\label{algline:7}
			\State $X^{k+1}_v = \gamma_v D_{j^*_{k+1},v}$ \qquad \qquad\qquad\qquad\qquad\quad\qquad\quad\,\#(rescaling)\label{algline:8}
			\EndFor 
			\If{$\mu_k \geq \frac{\norm{X^{k+1}-X^k}^2}{\norm{ \Aa(X^{k+1}-X^k) }^2}$} \qquad\qquad\qquad\qquad\quad\qquad\,\#(adaptive step-size shrinkage)
			\State $ \mu_k=\mu_k/2$ 
			\Else
			\State $k=k+1$
			\EndIf						
			\EndWhile	
			\State {$\overline \Theta_v \leftarrow$ look-up-table$\left(D_{j^*_{k+1},v}\right), \forall v$}\\
			\Return reconstructed MRF image $X^{k+1}$, parameter maps $\overline\Theta$, proton density $\gamma$
		\end{algorithmic}
		\caption{\label{alg:Coverblip} CoverBLIP$(Y,\Tt(D),\Aa,\mu)$} 
	\end{algorithm}
	

\subsection{CoverBLIP algorithm}
\label{sec:coverblip}
Approximation plays a key role in accelerating the nearest  neighbour searches
and breaking the curse-of-dimensionality~\cite{indyk1998breakCOD}. 
Motivated by the low-dimensional (manifold) structures present in the MRF dictionary, we propose to accelerate iterative matched-filtering steps within the BLIP algorithm  by using cover tree's $(1+\epsilon)$-ANNS approximate searches. Algorithm~\ref{alg:Coverblip} outlines the proposed Cover tree BLoch Iterative Projection (CoverBLIP) procedure for accelerated MRF reconstruction. We replace the exact NNS step \eqref{eq:exactP} in the cone projection with the following approximation:
\eql{\label{eq:ANN1}
	D_{j^*_{k+1}} = \textbf{$(1+\epsilon)$-ANNS}\left( Z_v/\norm{Z_v}, \Tt(D), D_{j^*_{k}}\right) \quad \forall v=1,\ldots,n,
}
which uses Algorithm~\ref{alg:NN} for a given inaccuracy level $\epsilon\geq 0$.
We denote by $\Tt(D)$ the cover tree structure built for the normalized fingerprint dictionary. At each iteration CoverBLIP uses previously selected fingerprints (i.e. $D_{j^*_{k}}=X_v^k/\gamma_v$ for each voxel) to initialize the ANNS searches. This has two positive impacts: i) the search achieves further acceleration especially, close to the converging point of the algorithm, because with an initialization close to the ANNS solution the branch-and-bound procedure can effectively rule out many branches at higher levels of the  tree and thus keep the candidates set very small, and ii) the $(1+\epsilon)$-ANNS algorithm would produce non-expansive outputs i.e. $\forall v$ we have
\eql{ 
	\norm{Z_v/\norm{Z_v} - D_{j^*_{k+1}}} \leq \norm{Z_v/\norm{Z_v} - D_{j^*_{k}}}, 
} 
which as will be discussed in the next part it is a key property to guarantee the monotone convergence of CoverBLIP. Note that we feed the search algorithm with the normalized gradient updates $Z_v/\norm{Z_v}$. Since dictionary atoms are normalized the search outcome is invariant with respect to the query rescaling, however from the complexity perspective one would gain in computation time by searching a query within a closer range to datasets' hypersphere. We also observed in our experiments that this trick leads to better accelerations. 

Convergence is tied to a proper choice of the step-size sequence. We follow the adaptive scheme proposed in~\cite{NIHT} which starts from a large initial step size and shrinks this choice by a division factor $\zeta>1$ e.g., half of the previous step size by setting $\zeta=2$ , until meeting the following criteria at each iteration $k$:
\eql{\label{eq:steprule}
	\mu_k < \frac{\norm{X^{k+1}-X^k}^2}{\norm{ \Aa(X^{k+1}-X^k) }^2}
		}
where again,  $\Aa(.) :=P_{\Omega} FS(.)$ is the shorthand for the forward operator. This condition is another important ingredient to guarantee the convergence of CoverBLIP iterations, which supported by some extra assumptions will also imply a robust reconstruction i.e. near global convergence. We will discuss this point in further details in the next section. 
After the first iteration we can also use the following energy ratio between measurements and our first estimation i.e. $\kappa={\norm{Y}}/{\norm{\Aa(X^1)}}$
in order to rescale the first iteration $X^1\leftarrow \kappa X^1$ and set an appropriate range (e.g. large enough) for the initial step size $\mu\leftarrow \kappa \mu$. 


When applicable \textemdash and with a possible compromise in the accuracy \textemdash a temporal subspace compression similar as explained in Section~\ref{seq:mrfdimchallenge} can
be optionally included to further shrink dimensions of $Z_v, X_v,D_j$ across the dominant SVD components $V_s\in \CC^{L\times s}$
of the MRF dictionary. 
In this case one has to build a cover tree structure for the normalized dimension-reduced dictionary $\widetilde D$, update the gradient step in Algorithm~\ref{alg:Coverblip} by the expression \eqref{eq:gradupdate}, and for the step-size  
expression~\eqref{eq:steprule} would change to
\eql{
	\mu < \frac{\norm{\widetilde X^{k+1}-\widetilde X^k}^2}{\norm{ \Aa(\widetilde X^{k+1}V_s^H-\widetilde X^kV_s^H) }^2}. 
} 
The updated gradient step might also introduce a compromise between cheaper distance evaluations 
during the search steps (i.e. in $\CC^s$ rather than $\CC^L$) and a computation overhead due to applying iteratively compression and decompression, as previously highlighted in Section~\ref{seq:mrfdimchallenge}. 

The approximate projection step presented in Algorithm~\ref{alg:Coverblip} (i.e. lines~\ref{algline:6} and~\ref{algline:7}) assumes that proton densities are real and positive valued quantities. A phase-alignment heuristic similar to~\cite{AIRMRF} can be used to extend this framework to complex-valued proton densities. This approach approximates dictionary atoms with fingerprints having constant complex phases across temporal domain. Complex angles corresponding to the first principal component i.e. $\widetilde D_{s=1}=V^H_{s=1}D$, are then used to  align dictionary atoms. Similarly, at each iteration in line~\ref{algline:6} we align phases of the gradient update used for the search step; In our experiments we use the complex angles of the dominant compressed image i.e. $\text{angle}(\widetilde X_{s=1})$ for temporal phase-alignment. 
Empirical results applying this approach are demonstrated for our volunteer data experiments in Section~\ref{sec:expe_invivo}. 

\section{Convergence of CoverBLIP}\label{sec:proofs}
The analysis in this part covers the behaviour of a wide class of \emph{inexact} IPG algorithms for solving linear inverse problems where the forward operator $\Aa$ and the set $\Cc$ of signal model could be regarded in general forms and not necessary customized for the MRF recovery problem. 

A previous work \cite{inexactipg-tit} studied the stability of the inexact IPG algorithms with respect to several forms of approximations on gradient and projection updates. Here we focus on iterative algorithms that use the following notion of relative approximate projection step i.e. for an $\epsilon\geq 0$ we define
\eql{\label{eq:eproj}
	\widetilde \Pp_\Cc^\epsilon(x) \in \left\{u\in \Cc: \norm{u-x}\leq (1+\epsilon) \inf_{u'\in\Cc}\norm{u'-x}\right\}.
	}
{\exmp{Following Definition~\ref{def:eNN}, the $(1+\epsilon)$-ANNS search algorithm is an approximate projection of type~\eqref{eq:eproj} onto a discrete set of points $\Cc:=D$ in a dataset e.g. a signal model which is used for data-driven inverse problems~\cite{inexactipg-tit}.} }
{\exmp{
Notably for projection onto $\Cc:=\text{cone}(D)$, if we replace the exact search step in~\eqref{eq:exactP} with an approximate $(1+\epsilon)$-ANNS search, we obtain an approximate cone projection $\Pp^\epsilon_{\text{cone}(D)}(.)$ satisfying definition \eqref{eq:eproj}. Steps~\ref{algline:6} to~\ref{algline:8} in CoverBLIP Algorithm~\ref{alg:Coverblip} are indeed implementing such an approximate projection onto the cone associated with the MR fingerprints using fast cover tree searches.}}

The corresponding inexact IPG iterations, including the CoverBLIP algorithm as a particular case, are as follows:
\eql{\label{eq:inIP2}
	X^{k+1} = \Pp^\epsilon_\Cc \left(X^k - \mu_k \Aa^H\left(\Aa(X^k)-Y \right)\right), 
}

We now follow this section by discussing two types of guarantees for the inexact IPG. The first type makes an embedding assumption on $(\Aa,\Cc)$ and provides a robust signal recovery result 
which in turn implies an interesting near global convergence guarantee for arbitrary signal models $\Cc$ including the non-convex conic constraints in the MRF problem. The second form of our analysis does not make an embedding assumption and only relies on an adaptive step-size scheme to ensure criteria~\eqref{eq:steprule} holds and guarantees local convergence of the inexact IPG algorithm. 

The following embedding assumption plays a critical role in our stable signal recovery result~\cite{BW:manifold,Blumen,inexactipg-tit}:
{\defn{\label{def:bilip} A forward operator $\Aa$ is bi-Lipschitz with respect to a set $\Cc$, if $\forall x,x'\in \Cc$ there exists constants $0<\alpha\leq \beta$ such that
\eql{\label{eq:bilip}
			\alpha \norm{x-x'}^2 \leq \norm{\Aa(x-x')}^2\leq \beta \norm{x-x'}^2.  }	
	}}
Equipped with this notion the following result states that when we have a good measurement consistency i.e. when $\min_{X\in\Cc}\norm{Y-\Aa(X)}$ is small,
then a near global convergence could be achieved using inexact iterations~\cite{inexactipg-tit,CoverBLIP-MLSP}:
 
{\thm{\label{th:inexactLS1} Assume $(\Aa,\Cc)$ is bi-Lipschitz and that for a given $\epsilon\geq 0$ and some constant $\delta \in [0,1)$ it holds
\eq{
\sqrt{\epsilon+\epsilon^2}\leq \delta \sqrt{\mmx}/{\vertiii{\Aa}} \qandq \MM < (2-2\delta+\delta^2) \mmx,}
where $\vertiii{\Aa}$ denotes the spectral norm of $\Aa$. 
Set the step size $\mu_k=\mu, \forall k$ such that \eq{\left((2-2\delta+\delta^2) \mmx\right)^{-1}<\mu\leq\MM^{-1}.} The sequence generated by Algorithm~\eqref{eq:inIP2} obeys the following bound:
		\eql{\label{eq:linconv}
			\norm{X^{k}-\xgt}\leq  \rho^k \norm{\xgt} + 
			\frac{\kappa_w}{1-\rho}w
		}
		where $\xgt=\argmin_{X\in \Cc}\norm{Y-\Aa(X)}$, $w=\norm{Y-\Aa (\xgt)}$ and
		\begin{align*}
		\rho=\sqrt{\frac{1}{\mu \mmx} -1}+ \delta, \quad
		\kappa_w= 2\frac{\sqrt{\MM}}{\mmx}+\sqrt\mu\delta.	
		\end{align*}				}} 	
{\rem{Theorem~\ref{th:inexactLS1} guarantees a linear convergence behaviour for inexact iterations. As a result after a finite $K=O(\log(\tau^{-1}))$ number of iterations Algorithm \eqref{eq:inIP2} achieves the solution accuracy $\norm{X^K-\xgt}=O(w)+\tau$ for any $\tau>0$.}}
{\rem{Under a properly conditioned bi-Lipschitz embedding as  assumed in~Theorem~\ref{th:inexactLS1} the inexact algorithm achieves a solution accuracy comparable to that of the exact IPG algorithm. By increasing $\epsilon>0$ we require better embedding conditions as compared to the exact iterations (i.e. the case  $\epsilon,\delta=0$). Although, increasing $\epsilon$ slows down the rate $\rho$ of linear convergence, it could facilitate significantly cheaper computations per iteration. 
In other words, approximation trades-off against the embedding conditions, rate of convergence and computation time, but not against the order of the solution accuracy.
}}
\newline

The following proposition (see the proof in Section~\ref{sec:proof0} of the supplementary materials) says that by using the adaptive shrinkage scheme described in the previous part we can find a good step size in a finite (logarithmic) number of sub-iterations:
{\prop{\label{prop:stepbound}Following the iterative step-size shrinkage scheme with the initial size $\mu$ and division factor $\zeta>1$, the chosen step size $\mu_k, \forall k$ meets the criteria~\eqref{eq:steprule} and satisfies the following bound:
		\eql{\label{eq:stepbound}
			(\zeta\MM)^{-1}\leq\mu_k\leq \mmx^{-1}. }  
in a finite number $\left \lceil \log_\zeta(\MM \mu) \right \rceil+1$ of iterations.}}
\newline 

The following theorem establishes a stable reconstruction guarantee (i.e. near global convergence) for the inexact IPG algorithm by using the adaptive step size shrinkage scheme:
{\thm{\label{th:inexactLS2} Assume $(\Aa,\Cc)$ is bi-Lipschitz, and that for given $\epsilon\geq 0$, $\zeta>1$ and some constant $\delta \in [0,1)$ it holds
\eq{
\sqrt{\epsilon+\epsilon^2}\leq \delta \sqrt{\mmx}/{\vertiii{\Aa}} \qandq \zeta \MM < (2-2\delta+\delta^2) \mmx.
}
Following the adaptive step-size scheme with shrinkage factor $\zeta$, the sequence generated by Algorithm~\eqref{eq:inIP2} obeys the error bound~\eqref{eq:linconv} 
where, 
		\begin{align*}
		\rho=\sqrt{\frac{\zeta\MM}{\mmx} -1}+ \delta, \quad
		\kappa_w= 2\frac{\sqrt{\MM}}{\mmx}+\frac{\delta}{\sqrt\mmx}.	
		\end{align*} 
}}
The proof architecture is similar to the proof of~\cite[Theorem~2]{inexactipg-tit}, however, this result does not a priori assume $\mu_k\leq 1/\MM$ as there or in Theorem~\ref{th:inexactLS1} of this paper. 
For the sake of completeness we provide detailed proof of Theorem~\ref{th:inexactLS2} in the supplementary materials Section~\ref{seq:proof1}  

{\rem{Without an a-priori knowledge of the embedding constants, the inexact IPG algorithm with adaptive step-size  exhibits a similar linear convergence behaviour towards the global minima as in Theorem~\ref{th:inexactLS1}. The closer $\zeta$ is chosen to one,  the embedding condition and the rate of convergence become more comparable to Theorem~\ref{th:inexactLS1}, however at the increased cost of more shrinkage sub-iterations.}}
{\rem{Theorem~\ref{th:inexactLS1} generalizes results in \cite{NIHT} in two ways: i) the set $\Cc$ of constraints are general and not restricted to sparse signals, and ii) results here establish robustness against inexact projection updates. Notably when no approximation is used $\epsilon,\delta =0$, Theorem~\ref{th:inexactLS1} relaxes the embedding conditions in \cite[Theorem 3]{NIHT} which required $\zeta\MM<8/7 \mmx$.
}}
\newline

Finally, we consider a general convergence result which holds even in the absence of the bi-Lipschitz embedding assumption. We additionally assume that the approximate projection produces non-expansive updates with respect to the previous iterations i.e. $\forall k$ and gradient updates $Z^k := X^k - \mu_k \Aa^H\left(\Aa(X^k)-Y \right)$ it holds :
\eql{\label{eq:nonexpansive}
\norm{\pp_\Cc(Z^k) - Z^k}\leq \norm{X^k - Z^k}.
}
{\exmp{The $(1+\epsilon)$-ANNS update~\eqref{eq:ANN1} in CoverBLIP and the associated approximate cone projection satisfy the non-expansiveness property~\eqref{eq:nonexpansive}, thanks to initializing the search algorithm with previous iteration.\footnote{In general one could easily incorporate property~\eqref{eq:nonexpansive} by the following update in Algorithm~\eqref{eq:inIP2}:
	$X^{k+1}=\argmin_{u\in \{\pp_\Cc(Z^k),X^k\}} \norm{u-Z^k}$.}
}}
\newline
\newline
The following result guarantees monotone convergence of Algorithm~\eqref{eq:inIP2} since the cost function $\norm{Y-\Aa(X)}\geq 0$ is lower bounded. The proof is provided in the supplementary materials Section~\ref{sec:proof2}:
{\thm{\label{th:inexactmonotone}Assume the approximate projections are non-expansive and the step-size satisfies~\eqref{eq:steprule}.  Algorithm~\eqref{eq:inIP2} produces a non-increasing and convergent sequence $\norm{Y-\Aa(X^k)}$.}}
\newline

%
Note that determining the bi-Lipschitz conditioning i.e. constants $\alpha$ and $\beta$ and hence an admissible interval for choosing the step-size (as suggested in Theorem~\ref{th:inexactLS1}) is a combinatorial problem in general. For a certain class of \emph{random} sampling schemes used in compressed sensing theory however it is possible to derive those constants with high probability, see e.g.~\cite{vershynin-randommatrix, rauhut-randommatrix,JLCS}.  Applied to the MRF problem, it has been shown in \cite[Theorem 1]{BLIPsiam} that if sampling patterns $\Omega_t$ sub-select uniformly at random large enough number of rows (or columns) of the k-space \textemdash a sampling protocol referred to as the random Echo Planar Imaging (EPI) \textemdash then the resulting forward model $\Aa$ is bi-Lipschitz, and a fixed choice of step-size  equal to the compression factor
\eq{\mu_k= n/m, \quad \forall k}
 guarantees stable reconstruction. Randomized acquisition schemes are however not currently popular in practical MRF setups, leading to pronounce more the importance of theorems~\ref{th:inexactLS2} and~\ref{th:inexactmonotone}. In Theorem~\ref{th:inexactLS2} one does not need to explicitly obtain the bi-Lipschitz constants, however if the forward model happens to satisfy a proper embedding condition then the adaptive step-size scheme is determined to make a proper choice (within the corresponding admissible interval) which guarantees near global convergence. Otherwise in the absence of any embedding assumption Theorem~\ref{th:inexactmonotone} ensures monotone convergence of the non-convex projected iterations i.e. the stability of the algorithm.


%% file: sections/expe.tex
\section{Numerical experiments}
\label{sec:expe}

\begin{table*}[t!]
	\centering
	\scalebox{.74}{
		\begin{tabular}{clc}
			\toprule[0.2em]
			Algorithm & Description  \\		
			\midrule[0.1em]
			BLIP & Iterative reconstruction \eqref{eq:blip} using exact brute-force searches for the matched-filtering~\cite{BLIPsiam} \\	
			\midrule[0.05em]
			Template Matching & Non iterative matched-filtering reconstruction using exact brute-force searches~\cite{MRF}\\
			(TM) & (i.e. the first iteration of BLIP)\\
			\midrule[0.05em]
			KDBLIP & Iterative reconstruction similar to~\cite{AIRMRF} using KD-tree's ANNS for the matched filtering \\
			& (Approximation level is controlled by the number of \emph{checks} which specifies the maximum leafs to\\ &\, visit during the search.
			Higher checks values give better search
			precision, but also take more time)\\
			\midrule[0.05em]
			CoverBLIP & Iterative reconstruction Algorithm~\ref{alg:Coverblip} using cover tree's $(1+\epsilon)$-ANNS for the matched filtering\\
			&(Approximation level is controlled by $\epsilon\geq 0$ which bounds the search precision according to \\ &\, Definition~\ref{def:eNN}. 
			Smaller $\epsilon$ would give better search precision, but also take more time).   \\
			\bottomrule[0.2em]
		\end{tabular}}
		\caption{Algorithms used for validations and comparisons.}\label{tab:algs}
	\end{table*}

In this section we evaluate and compare the performance of the proposed CoverBLIP algorithm against dictionary-based MRF reconstruction baselines listed in Table~\ref{tab:algs}. Experiments are conducted using MATLAB on a moderate desktop with 8 CPU-Cores and 32 GB RAM. For BLIP and TM algorithms the exact NNS is calculated using MATLAB's matrix product. KDBLIP iterations use 
randomized KD-tree searches implemented by the FLANN package \cite{flann}. 
Our CoverBLIP algorithm uses a parallel MATLAB interface to an existing implementation of the cover tree's $(1+\epsilon)$-ANNS in \cite{covertreecode}.\footnote{Implementations related to this work are available online at: \url{http://github.com/mgolbabaee/CoverBLIP}.} We do not believe this implementation is as optimized as that of the FLANN package for KD-tree searches and thus any reconstruction time comparisons (if not unfair) must take this point into account. 

\paragraph{Temporal subspace compression option} As discussed in Section~\ref{seq:mrfdimchallenge} all considered methods here can use a temporal compression option where the corresponding subspaces are the $s\leq L$ dominant SVD components of the fingerprint dictionary. This option has the advantage of reconstructing smaller objects i.e. MRF images, accelerated gradient updates i.e. forward and adjoint Fourier operations, and performing searches in low dimensional (ambient) space. The later is particularly beneficial for non-scalable search schemes such as KD-trees.

\paragraph{Datasets} Two sets of experiments are conducted: one using the synthetic Brainweb digital phantom with available Ground Truth (GT) maps~\cite{webbrain}, and the other using \emph{in-vivo} scans of a healthy volunteer's brain which appeared in the original work of Ma \emph{et al.}~\cite{MRF}. Both experiments use the Inversion Recovery (IR) Balanced SSFP acquisition sequence of length $L=1000$,  however with different flip angles and repetition times TR. The resulting  temporal signals from the IR-BSSFP sequence encode three NMR parameters $\Theta = \{T1,T2,B0\}$ i.e. the relaxation times and the off-resonance frequency. 
For each experiment and given FA and TR patterns a fingerprint dictionary is created as in~\cite{MRF} by solving discrete-time Bloch equations for combinations of the NMR parameters.

\begin{figure*}[t!]
	\centering
	\begin{minipage}{\textwidth}
		\centering
		\subfloat[Segments]
		{\includegraphics[height=4cm]{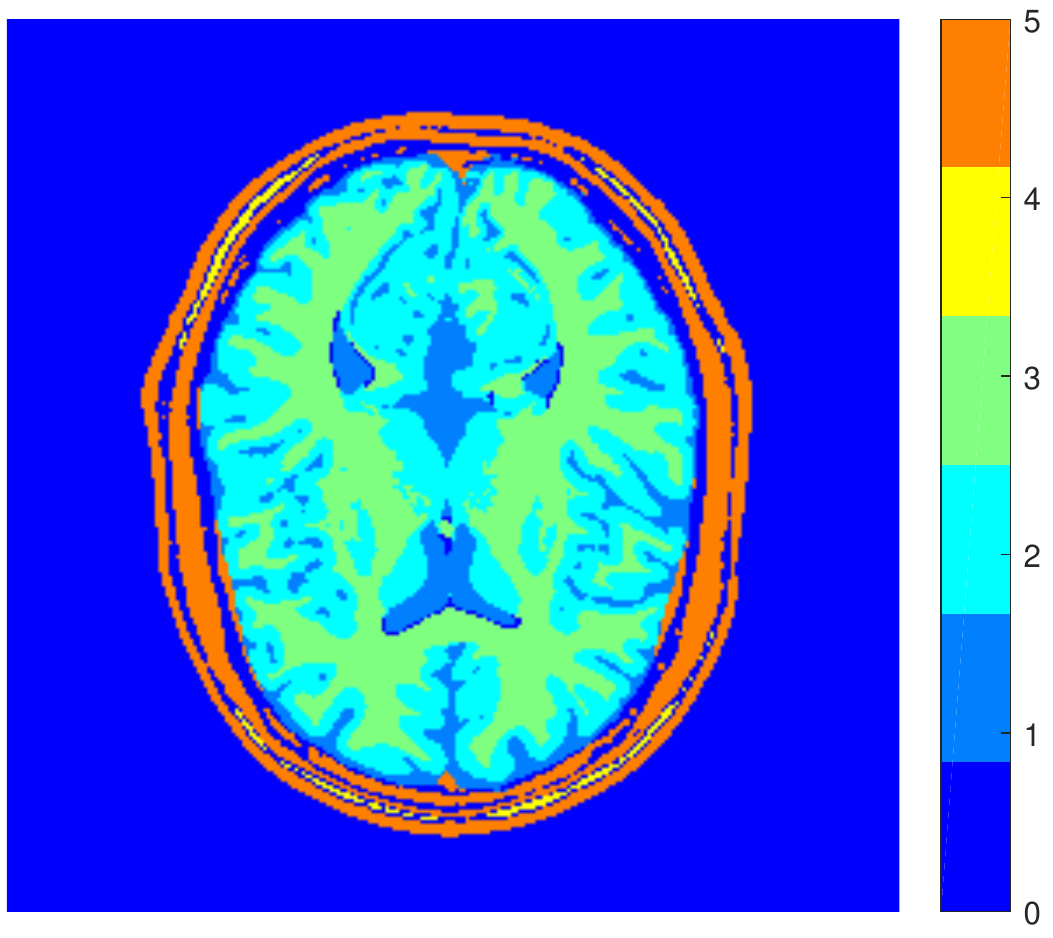} }
		\quad
		\subfloat[PD map]
		{\includegraphics[height=4cm]{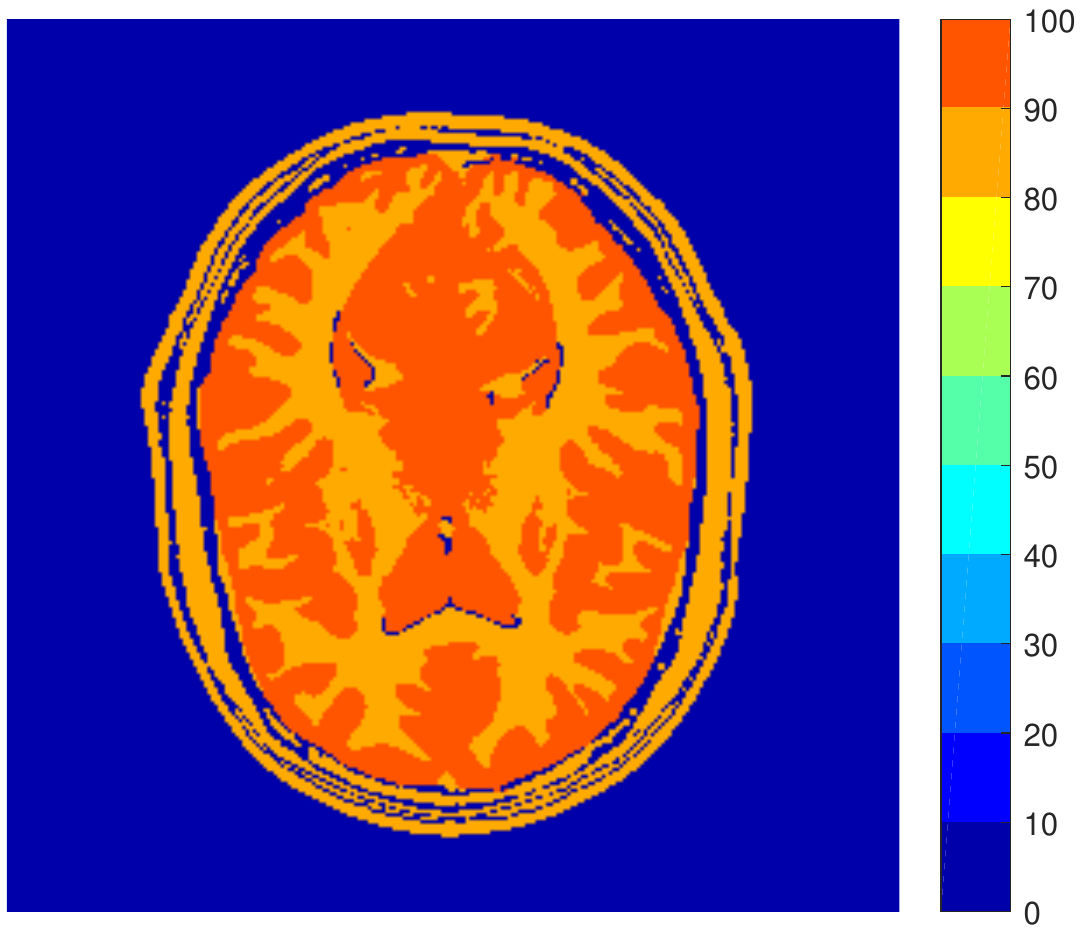} }	
		\caption{(a) The segmented Brainweb phantom coloured by index, and the corresponding parameters ($T1$ msec, $T2$ msec, $B0$ Hz) used for simulations: 0 = background, 1 = Cerebrospinal Fluid (5012, 512, -20), 2 = grey matter (1545, 83, -40), 3 = white matter (811, 77, -30), 4 = adipose (530, 77, 50), 5 = skin/muscle (1425, 41, 250), (b) the
			proton density (PD) map.
			\label{fig:numphantom}}
	\end{minipage}
\end{figure*}
\paragraph{Evaluation metrics} The normalized solution MSE (NMSE) is measured as $\frac{\|\hat{X}-X_0\|}{\|X_0\|}$
where $X_0,\hat{X}$ are the ground truth and the reconstructed MRF images, respectively. The NMR parameter estimation accuracies are measured e.g. for the $T1$ case, as follows:
\eq{
\text{$T1$ accuracy}:=1-\frac{1}{\Card(\Nn)}\Sigma_{v\in\Nn} \frac{|\hat T1(v)-T1(v)|}{T1(v)}
}
where $v$ is the number of voxels within a masked region $\Nn$ defining the object of interest. The mask is obtained by contouring the output proton density (PD) map of the brain and to remove empty voxels where the quantitative values are undefined. Here $T1(v)$ represents the ground truth $T1$ value for the $v$-th voxel and $\hat{T1}(v)$ is the corresponding estimated value.

\paragraph{Computational cost} To have a fair comparison between   computational complexities of the considered methods \textemdash and independent from how optimally they are implemented \textemdash we measure \emph{total search costs} in addition to the reconstruction times. The cost measures the total
number of computed pairwise distances (multiplied by the search dimension i.e. either $L$ or $s\leq L$ when using subspace compression) for performing the
NNS or ANNS steps within (iterative) matched-filtering until the algorithm converges. For all iterative methods the maximum number of iterations is set to $50$ and the algorithm is stopped earlier if the relative progress in minimizing the objective function of \eqref{eq:CS} (or \eqref{eq:CS1} when using subspace compression) is less than $10^{-6}$.

\begin{figure*}[t!]
	\centering
	\begin{minipage}{\textwidth}
		\centering
		\subfloat[Flip angles (FA)]
		{\includegraphics[width=.48\textwidth]{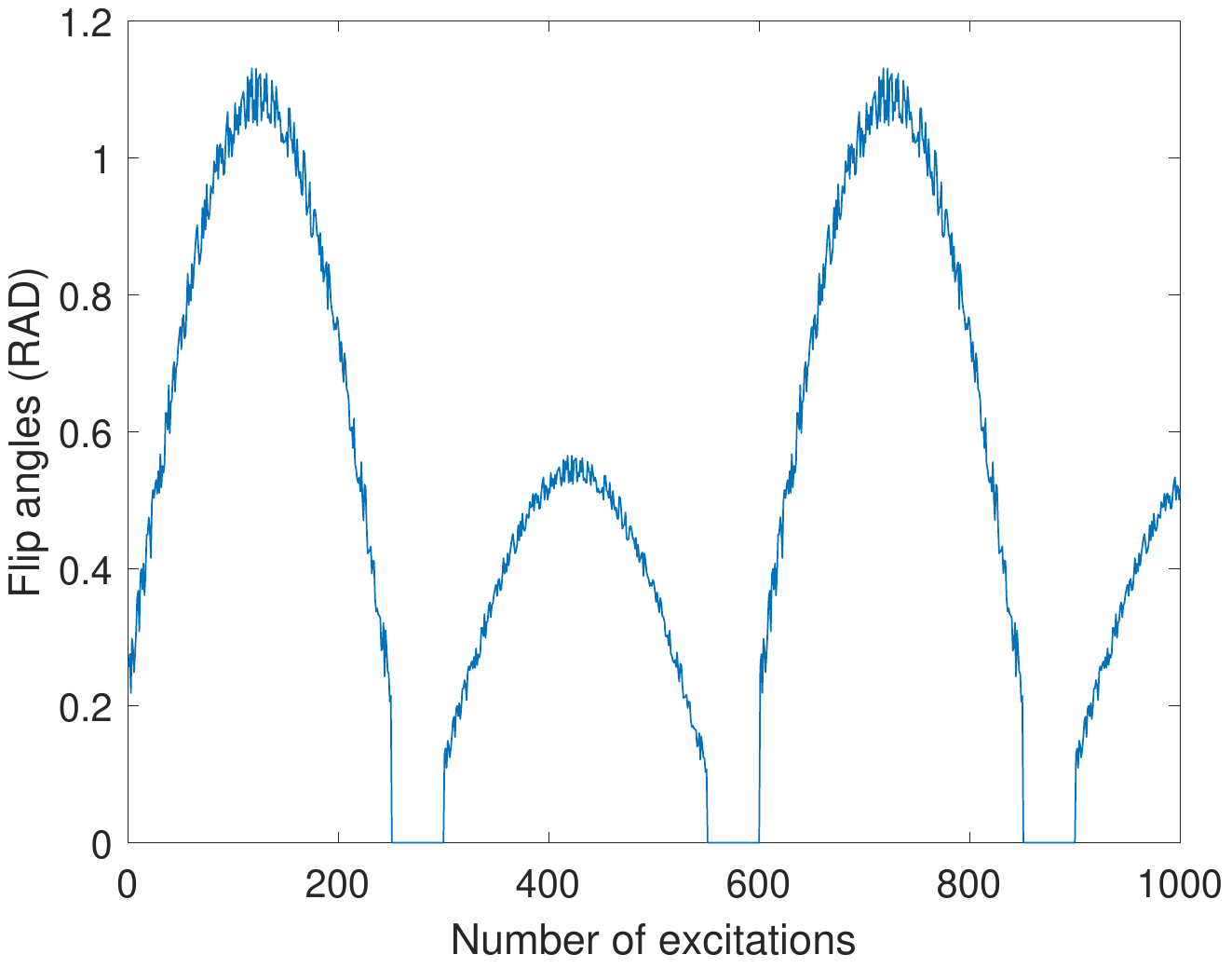} }
		\hfill
		\subfloat[$T2=100$ (msec), $B0=20$ (Hz)]
		{\includegraphics[width=.48\textwidth]{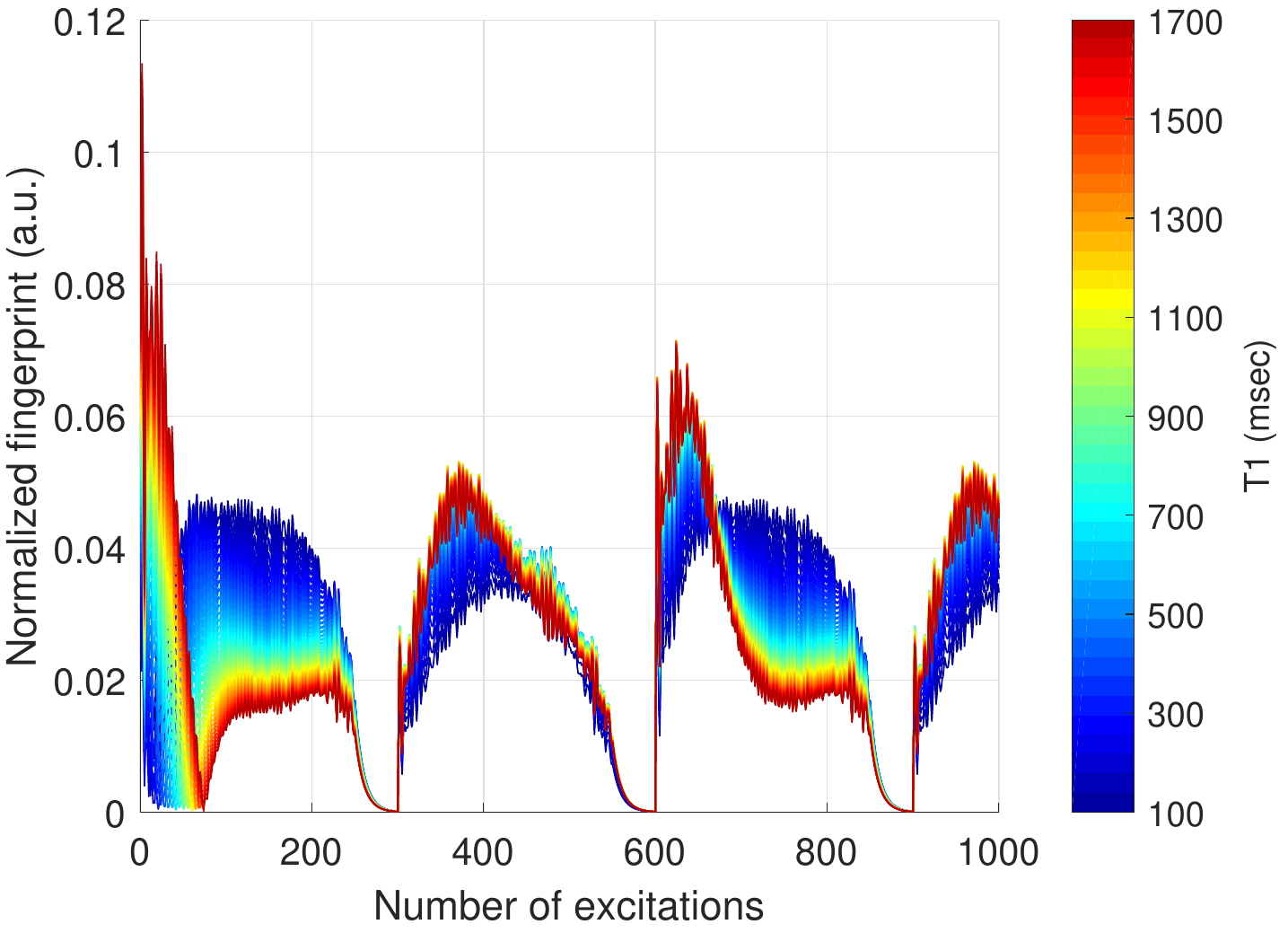} }
		\\
		\subfloat[$T1=1100$ (msec), $B0=20$ (Hz)]
		{\includegraphics[width=.48\textwidth]{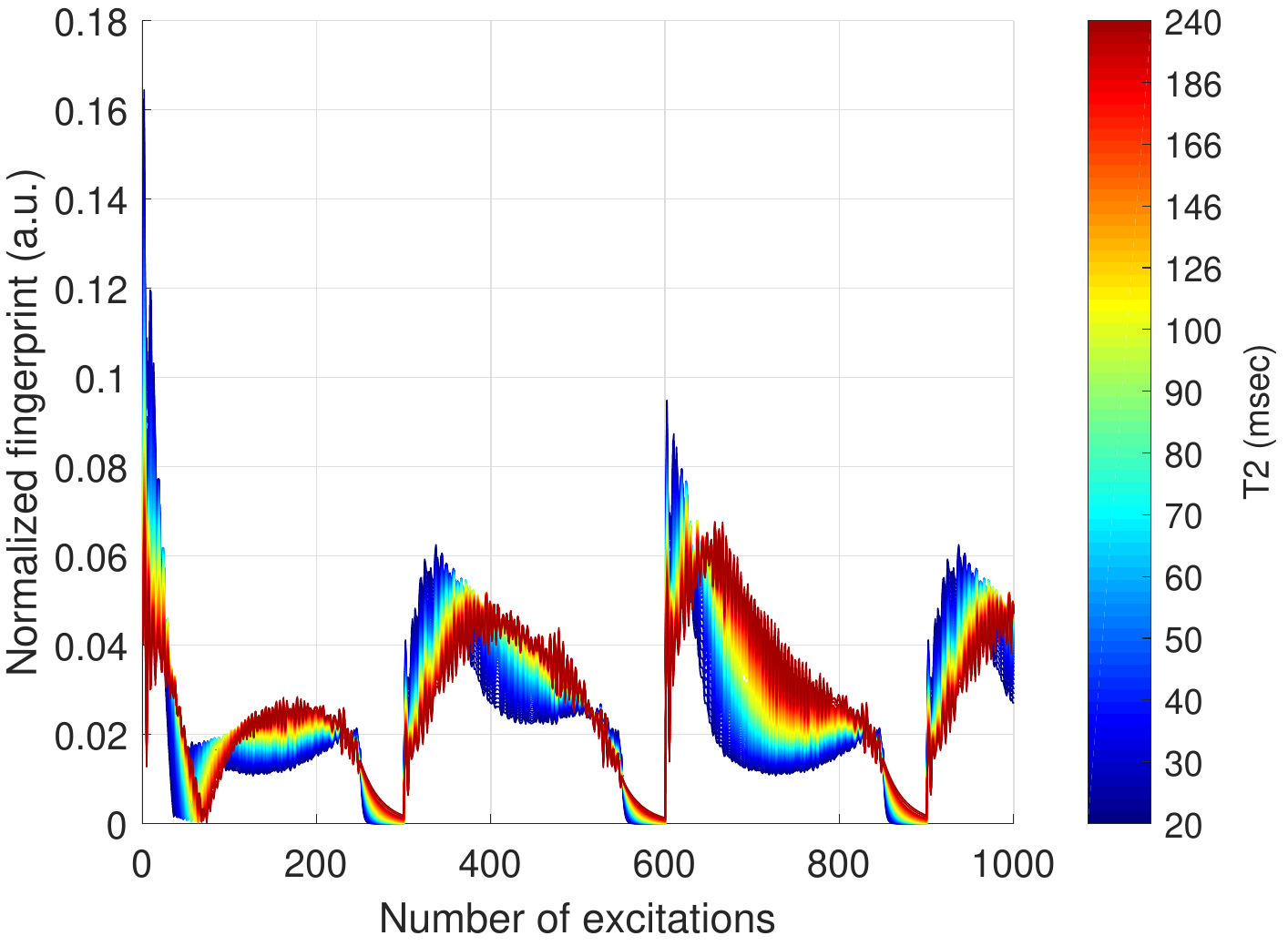} }	
		\hfill
		\subfloat[$T1=1100$ (msec), $T2=100$ (msec)]
		{\includegraphics[width=.48\textwidth]{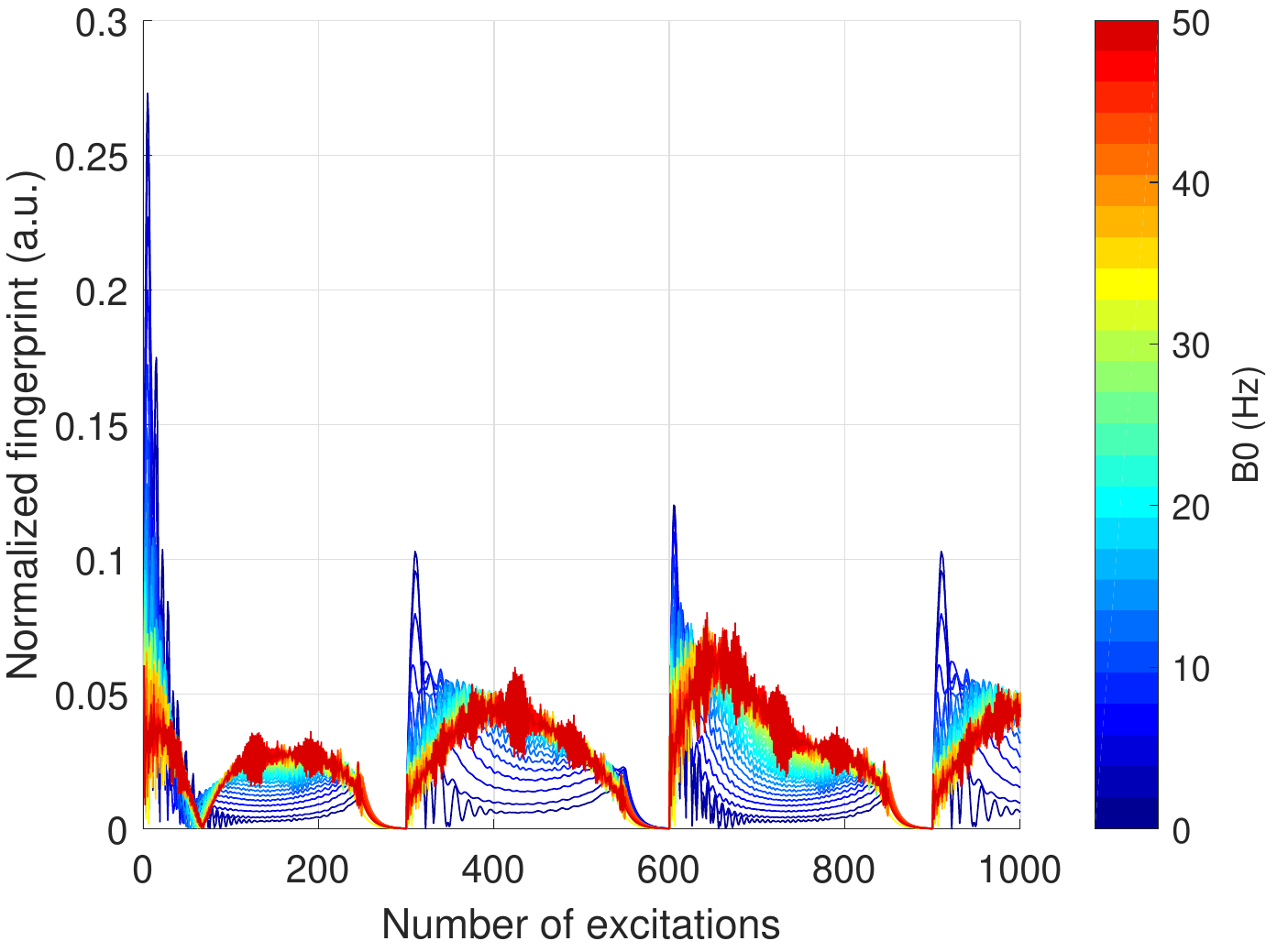} }		
		\caption{The IR-BSSFP dictionary generated from a set of pseudo-random FAs shown in Figure (a) and fixed $TR=10$ (msec). This dictionary encodes $T1$ and $T2$ relaxation times and off-resonance frequency $B0$. Figures (b)-(d) show the  magnitude of the complex fingerprints (i.e. dictionary atoms) for different parameter combinations. \label{fig:FAdict1}}
	\end{minipage}
\end{figure*}

\begin{figure*}[h!]
	\centering
	\begin{minipage}{\textwidth}
		\centering
		\subfloat
		{\includegraphics[width=.98\textwidth]{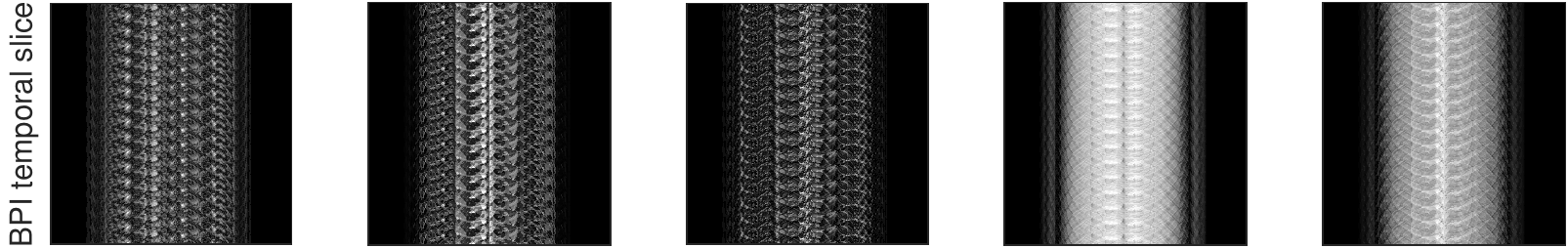} }
		\\
		\subfloat
		{\includegraphics[width=.98\textwidth]{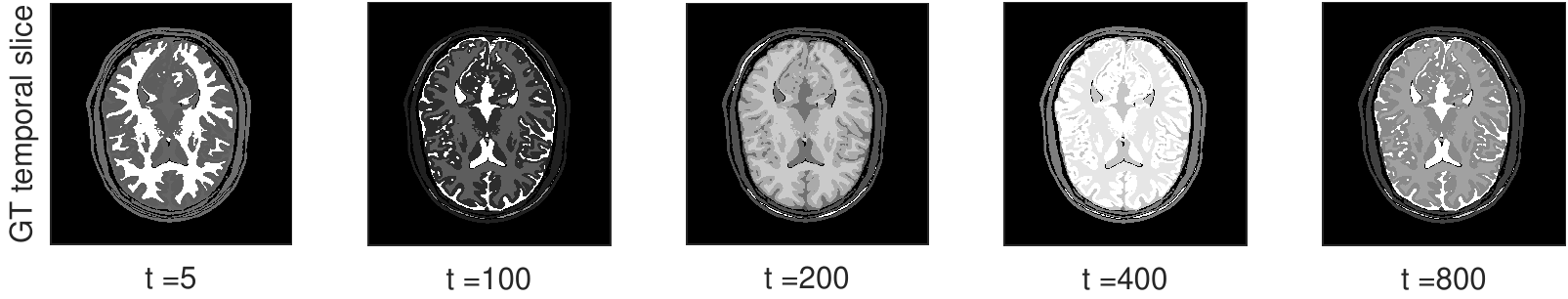} }	
		\caption{Ground truth MRF images generated from the Brainweb phantom (bottom row) and the highly aliased BPIs (top row)  $X'= \Aa^H(Y)$ across different time frames.
			\label{fig:BPI_synth}}
	\end{minipage}
\end{figure*}
\begin{table*}[t!]
	\centering
	\scalebox{.8}{
		\begin{tabular}{lccccccccc}
			\toprule[0.2em]
			Algorithm & checks/$\epsilon$ &  NMSE &  T1 acc. &T2 acc. & df acc. & search cost & search time& iter. & total runtime    \\
			& &  &  (\%) & (\%) & (\%) & &(sec)&  &  (sec)    \\
			\midrule[0.1em]
			& \multicolumn{8}{c}{Temporal compression $s=20$ }   \\
			\midrule[0.1em]
			BLIP&-& 1.563e-1& 94.2 & 85.3& 75.3& 1.11e13&1.08e3&14&1.11e3 \\
			\midrule[0.05em]			
			CoverBLIP&0.4& {1.295e-1}& {94.6}& {89.2}&{81.2}&{ 3.94e9}& {2.74e1}&14& {5.65e1} \\
			\midrule[0.05em]
			KDBLIP&256& 1.355-1& 94.2& 85.8& 79.6& 5.83e9 & 6.62e1&14&9.34e1 \\
			\midrule[0.1em]
			& \multicolumn{7}{c}{No temporal compression}   \\
			\midrule[0.1em]
			BLIP&-& 5.327e-3& {99.4} & {98.5}& {84.3}& 6.59e14&1.33e4 &20&1.34e4 \\
			\midrule[0.05em]
			CoverBLIP&0.4& {5.300e-3}& {99.4}& {98.5}& {84.3}& {4.86e11}& {3.45e2} &25& 4.93e2 \\
			\midrule[0.05em]
			KDBLIP&256& 1.727e-1& 92.7& 78.7& 67.4& 3.05e12& 4.49e2&38&6.96e2 \\
			\bottomrule[0.2em]
		\end{tabular}}
		\caption{Comparisons between iterative methods with/without using temporal subspace compression in terms of reconstruction NMSE, parameter estimation accuracy, search cost/time, number of iterations and total reconstruction time. 
		}\label{tab:phantom_comp}
	\end{table*}
\subsection{Numerical Brainweb phantom with multi-shot EPI acquisition}\label{sec:brainwebexpe}
In this part we compare the performance of different MRF recovery methods on a synthetic dataset. We use a $256\times256$ slice from the numerical Brainweb phantom with segments corresponding to the background and five different tissues each associated to a set of $\Theta=\{T1,T2,B0\}$ parameters (Figure~\ref{fig:numphantom}(a)). After an inversion pulse, a pseudo-random FA sequence\footnote{The FA pattern consists of half-sinusoidal
	curves with the period of 250 repetition times followed by no excitations for 50 repetition times between the cycles. The maximum flip
	angles alter between $60^\circ$ and $30^\circ$ in odd and even periods. A  zero-mean uniformly distributed noise of standard deviation 5 is added to the FAs.} with fixed $TR=10$ (msec) and $TE=TR/2$ is used for excitations.
Temporal Bloch responses of the phantom segments 
are amplified by a given PD map (Figure~\ref{fig:numphantom}(b)) and synthesize the whole ground truth MRF image $X_0$ of size $(n=256^2)\times (L=1000)$. The same excitation is used to generate a dictionary of $d=314160$ fingerprints corresponding to combinations of the quantized parameters 
$T1=[100:40:2000,2200:200:6000]$ (msec), $T2 = [20:2:100,110:4:200,220:20:600]$ (msec), $B0=[-250:40:-190,-50:2:50,190:40:250]$ (Hz). Figure~\ref{fig:FAdict1} illustrates the FA pattern and the resulting dictionary fingerprints used in this experiment.

For k-space sampling we simulate a similar protocol to the recently proposed multi-shot Echo Planar Imaging (EPI) for MRF acquisition~\cite{multishotEPI}. This protocol is based on a Cartesian grid Fourier sub-sampling where at each repetition time 16 out of 256 lines (with uniform spacing) from the k-space are simultaneously measured. In the next time frame the sixteen-shot sampling pattern $\Omega_t$ will be shifted by one line and so on. As a result we are dealing with reconstructing a 16x-fold undersampled data. We consider a single coil acquisition $S(X)=X$ and white Gaussian noise of 50 dB SNR added to the k-space measurements. Figure~\ref{fig:BPI_synth} illustrates the ground truth MRF images $X_0$ and the  highly aliased back-projected images (BPI) i.e. $X' = \Aa^H(Y)$ using this sampling protocol.

\begin{figure}
	\centering
	\begin{minipage}{\textwidth}
		\centering
		\subfloat
		{\includegraphics[width=.7\textwidth]{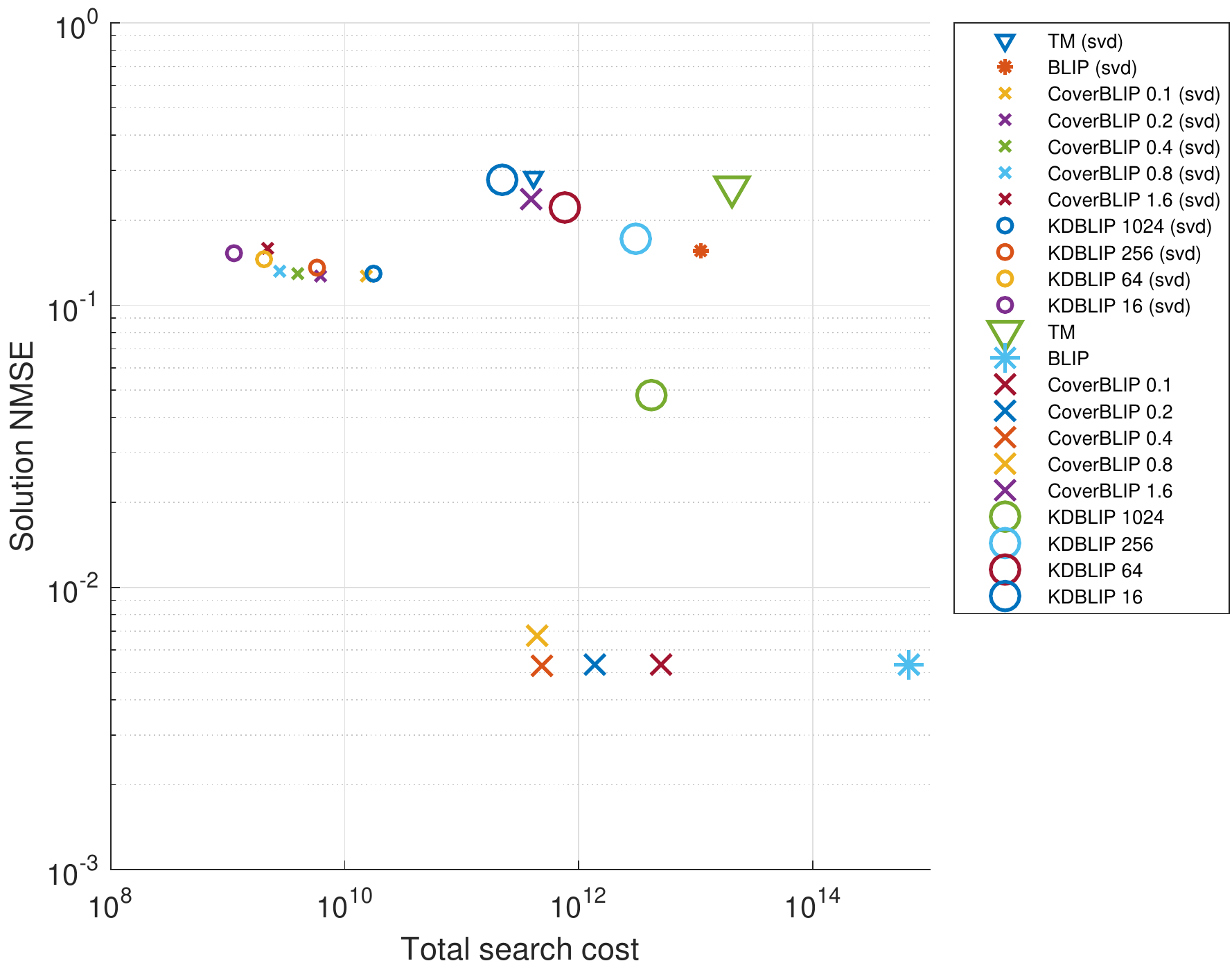} }
		\caption{Total search cost vs. solution accuracy of the non-iterative TM and iterative BLIP algorithms using brute-force searches, and inexact iterative algorithms CoverBLIP and KDBLIP using fast tree searches. Two scenarios of applying temporal SVD compression where $s=20$, and using no temporal compression are compared.
		\label{fig:compVSacc}}
	\end{minipage}
\end{figure}

	\begin{figure*}
		\centering
		\begin{minipage}{\textwidth}
			\centering
			\subfloat
			{\includegraphics[width=.98\textwidth]{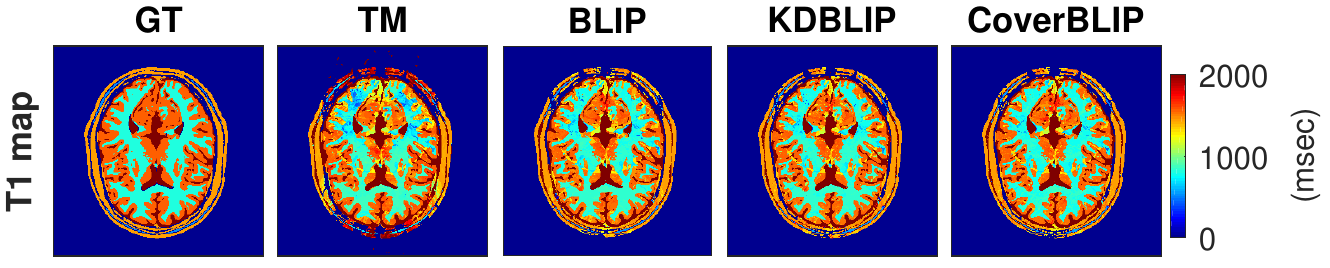} }
			\\
			\subfloat
			{\includegraphics[width=.98\textwidth]{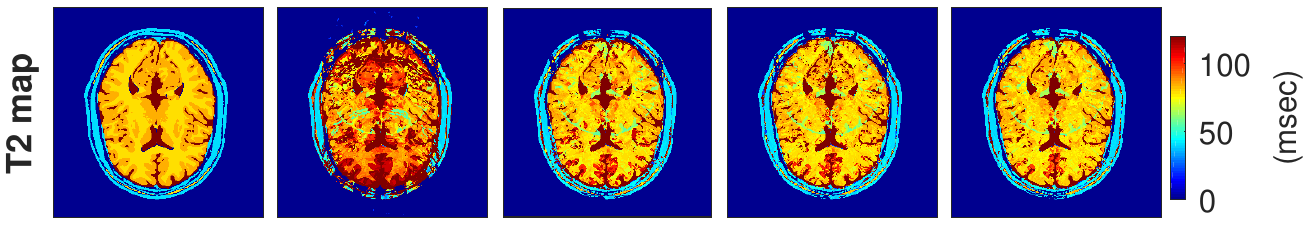} }
			\\
			\subfloat
			{\includegraphics[width=.98\textwidth]{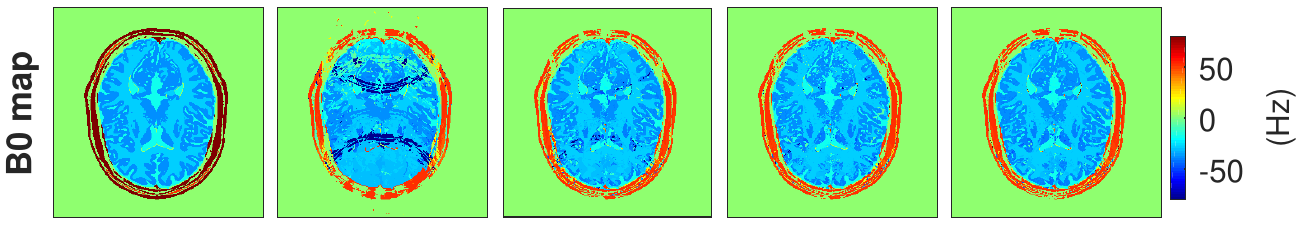} }
				\\
				(a) Temporal compression ($s=20$)
				\\
				\subfloat
				{\includegraphics[width=.82\textwidth]{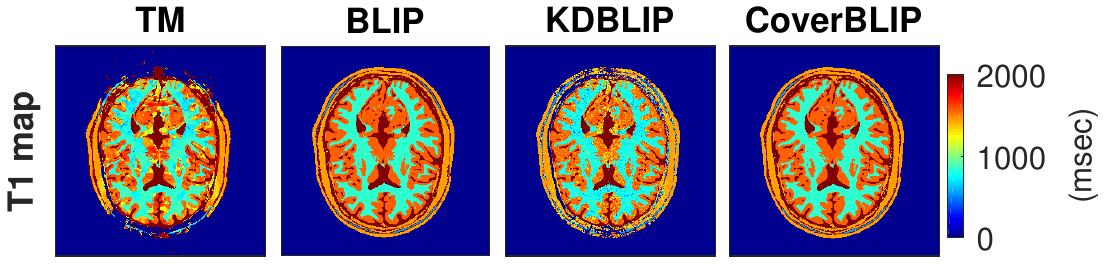} }
				\\
				\subfloat
				{\includegraphics[width=.82\textwidth]{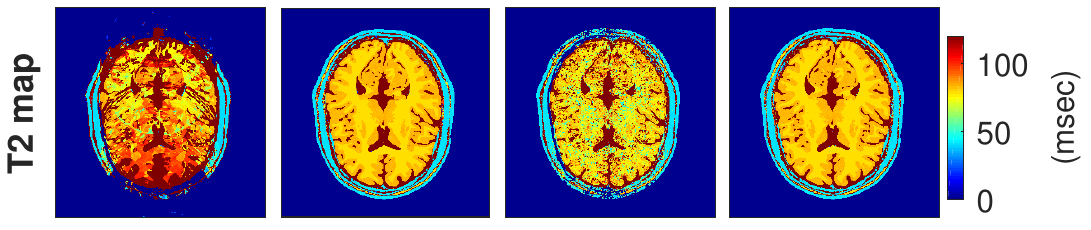} }
				\\
				\subfloat
				{\includegraphics[width=.82\textwidth]{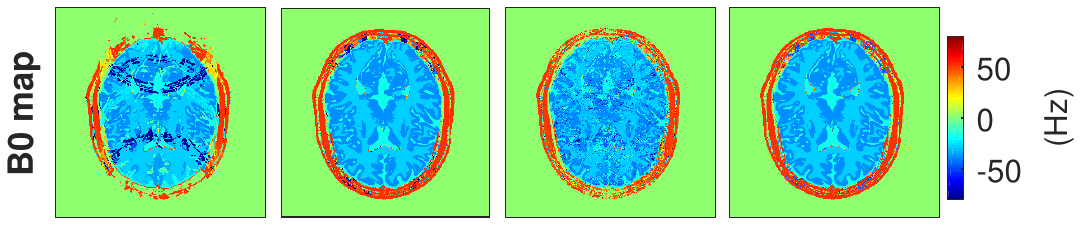} }		\\
							(b) No temporal compression
			\caption{The ground truth (GT) and reconstructed $T1$, $T2$ and $B0$ maps for the numerical Brainweb phantom acquired by the (simulated) multi-shot EPI protocol with x16-fold under-sampling. KDBLIP and CoverBLIP iterations use search accuracies checks = 256 and $\epsilon=0.4$, respectively. Figures (a) and (b) compare the estimated maps with/without using the SVD based temporal compression for all tested algorithms.\label{fig:phantom_maps}}
		\end{minipage}
	\end{figure*}

\subsubsection{Results}
We report reconstruction times, total search (projection) costs,  image reconstruction errors and parameter estimation accuracies in Figure~\ref{fig:compVSacc} and Table~\ref{eq:CS}. We also show the reconstructed parameter maps in Figure~\ref{fig:phantom_maps}. For the KDBLIP algorithm we vary the KD-tree's search accuracy level by choosing $\text{checks}=\{1024, 256, 64, 16\}$. For the CoverBLIP algorithm we also test different $(1+\epsilon)$-ANNS search approximations by choosing $\epsilon=\{0.1, 0.2, 0.4, 0.8, 1.6\}$.  We initialize the step-size of the iterative schemes by the compression factor $\mu=n/m$ which empirically turns out to satisfy criteria~\eqref{eq:steprule} in most iterations and requires one  shrinkage sub-iteration for the rest (see discussions in Section~\ref{sec:proofs}). 

Temporal SVD compression ($s=20$) accelerates the runtimes of all tested methods within 1-2 orders of magnitude (Table~\ref{tab:phantom_comp}), however such an aggressive compression leads to poor parameter reconstructions (Figure~\ref{fig:phantom_maps}(a)). 
Focusing on the non-compressed regime, we can see that Template Matching (TM) cannot achieve a good accuracy compared to the iterative methods (Figures~\ref{fig:compVSacc} and~\ref{fig:phantom_maps}(b)). The BLIP algorithm addresses this issue however at a high computational cost of iterating exact brute-force searches.
Note that since the multi-shot EPI acquisition uses a Cartesian sampling, $F$ in the forward model~\eqref{eq:CS} corresponds to a FFT operator with fast gradient updates. As a result, and as can be observed in Table~\ref{tab:phantom_comp}, projections (i.e. searches) dominate the runtimes of the iterative methods and thus accelerating this step would directly improve the total reconstruction time. CoverBLIP does so by using inexact cover tree searches (e.g. $\epsilon=0.4$) and achieves the best reconstruction time-accuracy (also search cost-accuracy) in all cases. Remarkably, CoverBLIP reports a similar accuracy to BLIP iterations however with 3 orders of magnitude less search cost and 27x-fold acceleration in the reconstruction time. Notably, the total cost of CoverBLIP inexact searches does not exceed that of a single stage brute-force search in TM (Figure~\ref{fig:compVSacc}). 

When using temporal compression\textemdash a favourable case for the KD-tree searches\textemdash KDBLIP with number of checks = 256 performs comparable to the CoverBLIP algorithm. However, for improving the overall estimation accuracy if we wish to not use subspace compression, then KDBLIP's time-accuracy performance fails to catch up with that of CoverBLIP.  Figure~\ref{fig:compVSacc} shows the gap between performances of these two algorithms caused by the non-scalability of the KD-tree searches. 
For instance CoverBLIP with $\epsilon=0.4$ outputs more accurate parameter maps (Figure~\ref{fig:phantom_maps}(b))  whilst reporting 6x less total search cost.

\subsection{In-vivo data with variable-density spiral acquisition}\label{sec:expe_invivo} 
In this part we evaluate reconstruction methods in Table~\ref{tab:algs} on \emph{in-vivo} MRF data acquired from a healthy volunteer using the IR-BSSFP sequence and the 1.5 T whole body Espree Siemens Healthcare scanner with 32-channel head receiver coil. This dataset was used in the seminal paper of Ma \emph{et al.}~\cite{MRF}, where FAs have a  pseudo-randomized (Perlin noise) pattern of length $L=1000$ and TRs are uniformly selected at random between 10.5 and 14 msec. At each time frame (repetition time) one interleaf of the variable-density spiral readout samples the k-space (see~\cite[Figure 1]{MRF} for the FA,TR and spiral readout patterns used in this experiment). The spiral trajectory $\Omega_t$ rotates by $7.5^\circ$ in the next time frame to sample different k-space locations and so on. The overall k-space undersampling factor is 48x folds and since a non-Cartesian readout pattern has been used, the operator $F$ in the forward model \eqref{eq:forward} is implemented using the non-uniform Fourier transform (NUFFT)~\cite{NUFFT}. 
Sensitivity maps (i.e. $S$ operator in \eqref{eq:forward}) are computed off-line from the acquired multi-coil data~\cite{cmap-adaptive}. For reconstruction a dictionary of $d=363624$ fingerprints is simulated for combinations of discrete parameters  
$T1=[100:20:2000,2300:300:5000]$ (msec), $T2 = [20:5:100,110:10:200,300:200:1900]$ (msec), and  $B0=[-250:20:-190,-50:1:50,190:20:250]$ (Hz). 
As a common practice used to precondition non-Cartesian MRF problems, we incorporate a density compensation scheme within the reconstruction pipeline to enable faster convergence (see more details in Section~\ref{sec:dcf} of the supplementary materials). With this update, we initialize the step-size by the compression factor $\mu=n/m$ similar to the Cartesian sampling. We empirically observe that this choice satisfies the criteria~\eqref{eq:steprule} for most of the iterations and for the rest one or two shrinkage sub-iterations suffices.
%
%
%
%

%

\begin{figure*}
	\centering
	\begin{minipage}{\textwidth}
		\centering
		\subfloat[T2 Map reconstructed by TM i.e. the first iteration of BLIP]
		{\includegraphics[height=4.5cm]{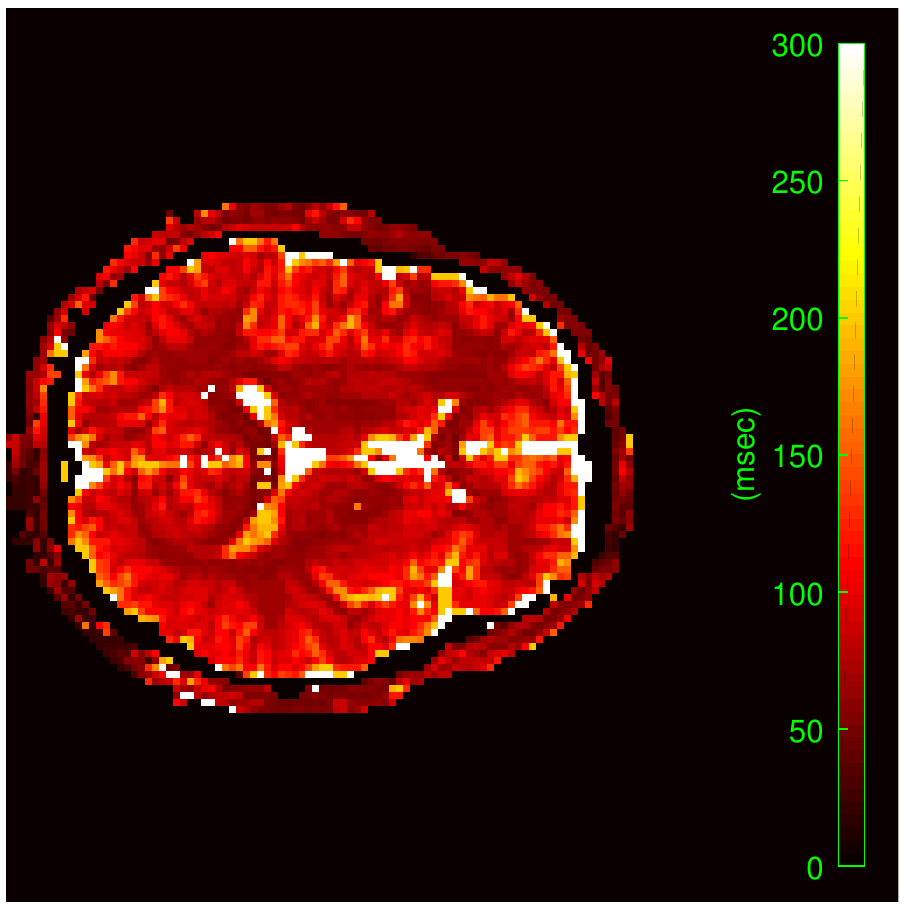} }
		\hfill
		\subfloat[T2 Map reconstructed by BLIP]
		{\includegraphics[height=4.5cm]{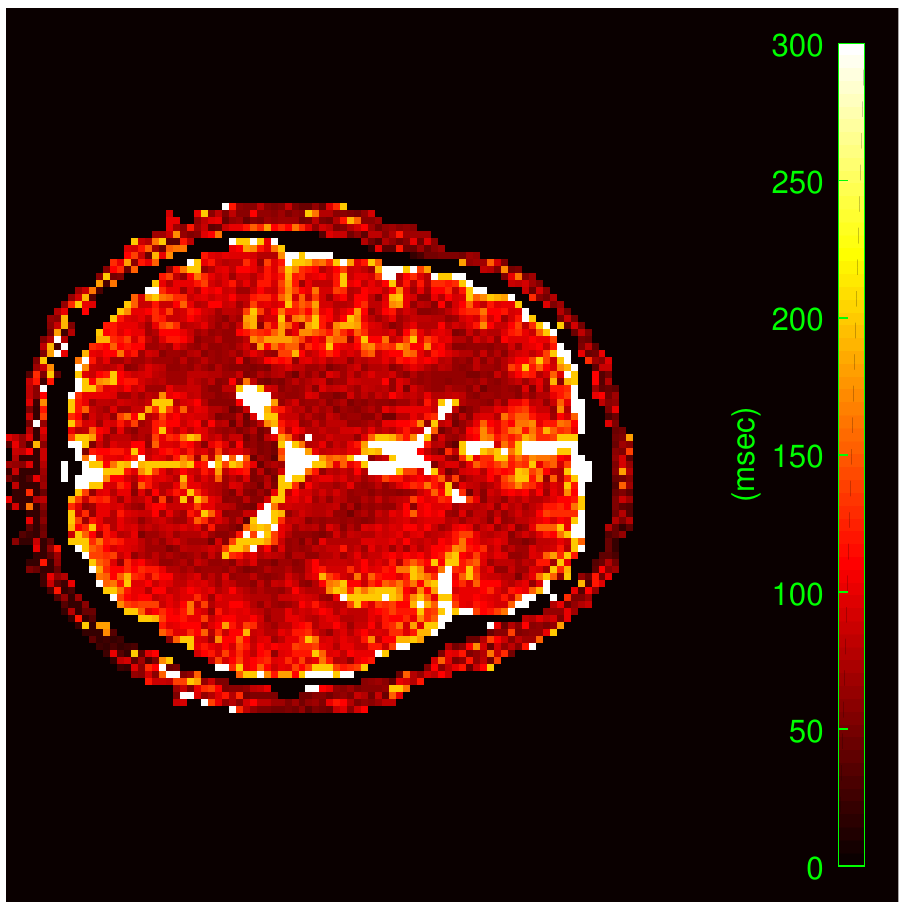} }
		\hfill			
		\subfloat[Fidelity error decay through iterations]
		{\includegraphics[height=4.5cm]{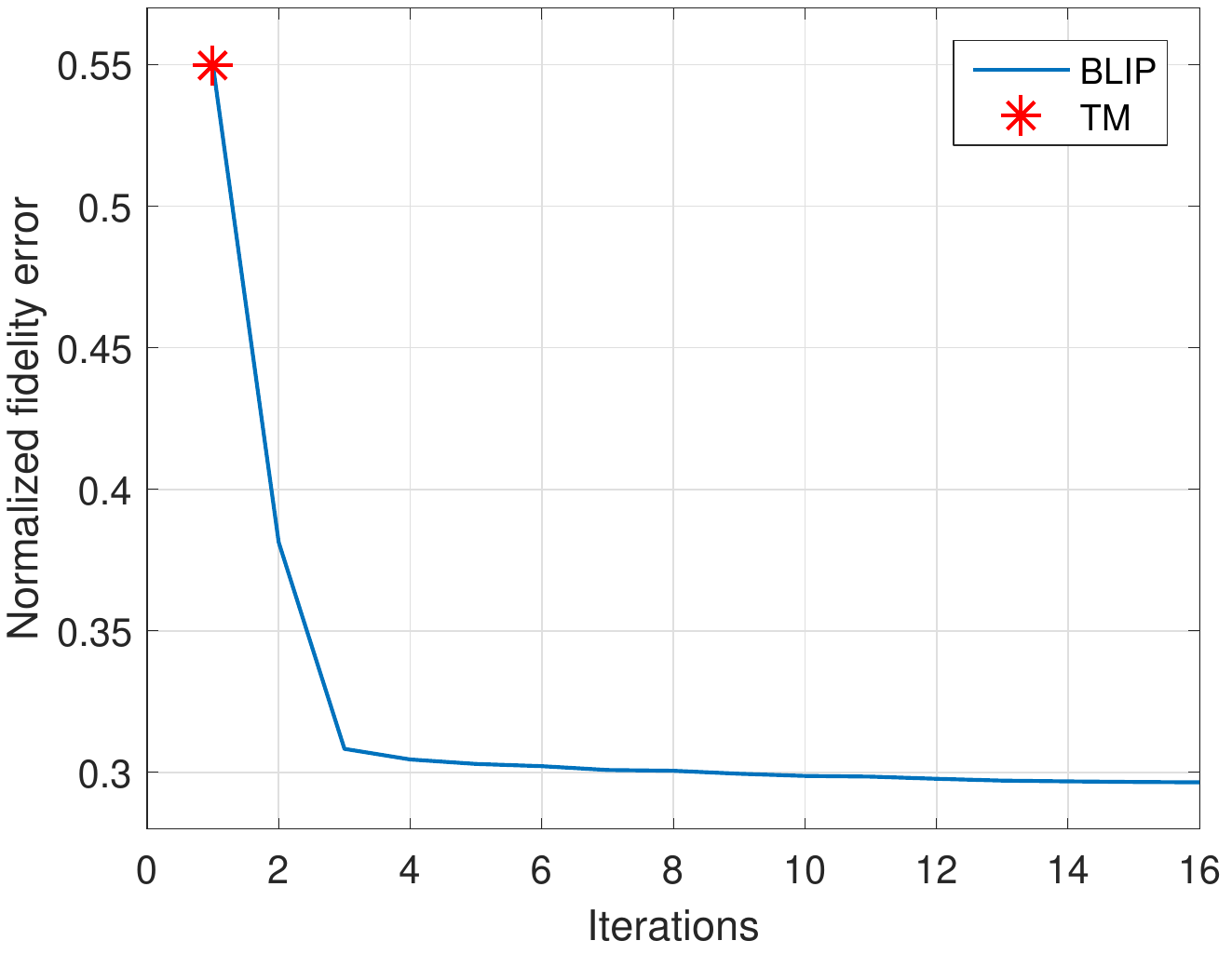} }	
		\caption{Reconstruction at $n=128\times128$ spatial resolution (32-coil data). Despite a better data consistency, the iterative scheme (BLIP) reports high-frequency artefacts  in the reconstructed maps indicating the lack of sufficient high-resolution information in an spiral readout.} \label{fig:highres}
	\end{minipage}
\end{figure*}


\begin{figure*}
	\centering
	\begin{minipage}{\textwidth}
		\centering
		\subfloat
		{\includegraphics[width=.95\textwidth]{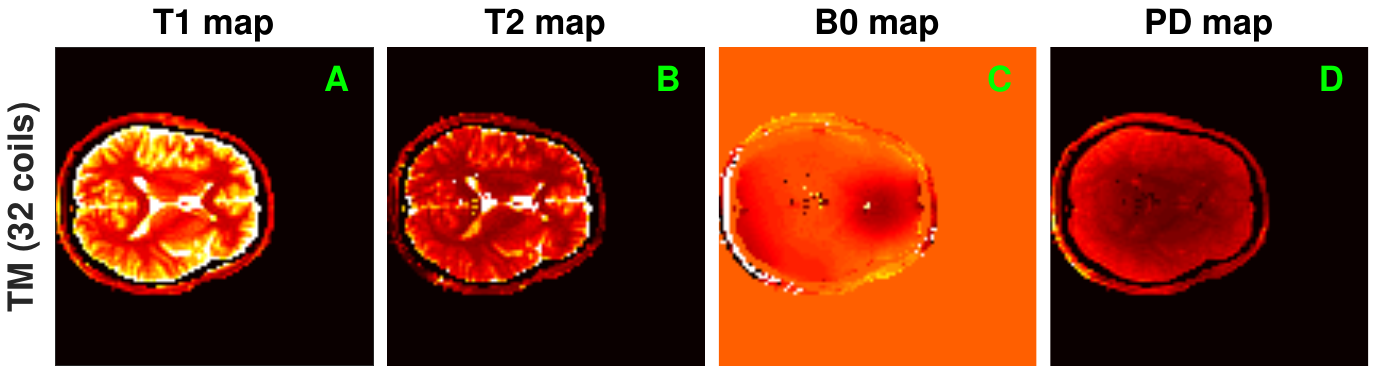} }
		\\
		\subfloat
		{\includegraphics[width=.95\textwidth]{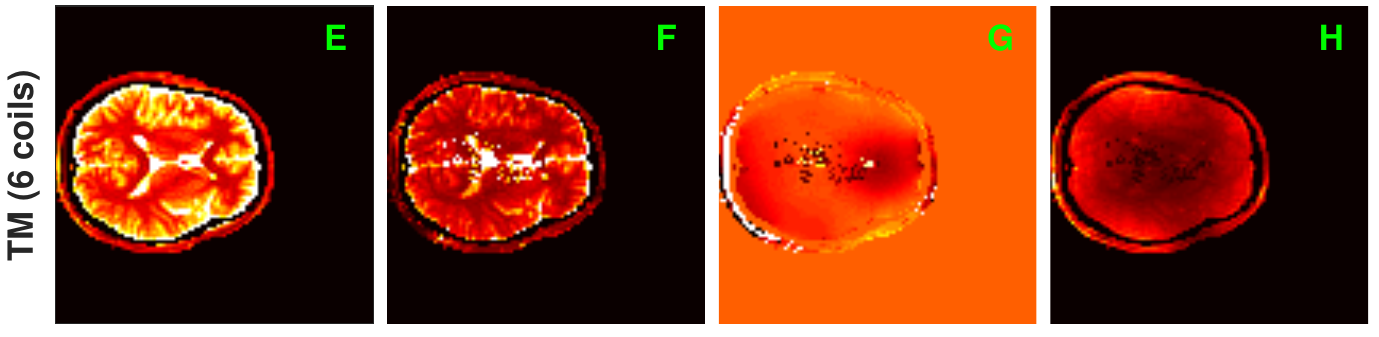} }
		\\
		\subfloat
		{\includegraphics[width=.95\textwidth]{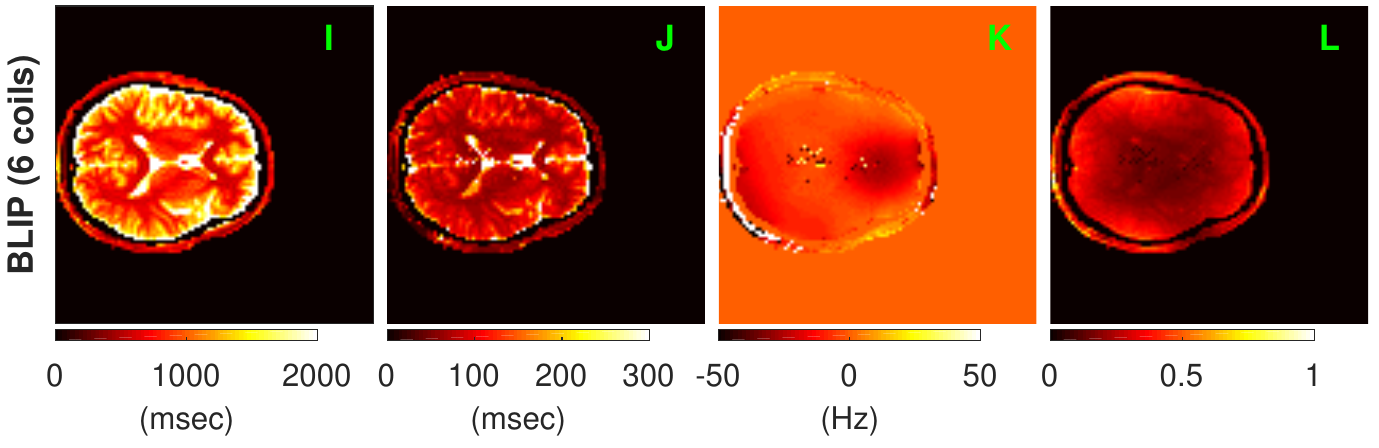} }		
		\caption{Recovered $T1,T2,B0$ and PD maps from the 32-coil data using TM algorithm (A-D), and those reconstructed from only 6-coil data by using TM (E-H) and iterative BLIP (I-L) algorithms.
				\label{fig:coilcompare}}
	\end{minipage}
\end{figure*}

\begin{figure*}
	\centering
	\begin{minipage}{\textwidth}
		\centering
		\subfloat
		{\includegraphics[width=.95\textwidth]{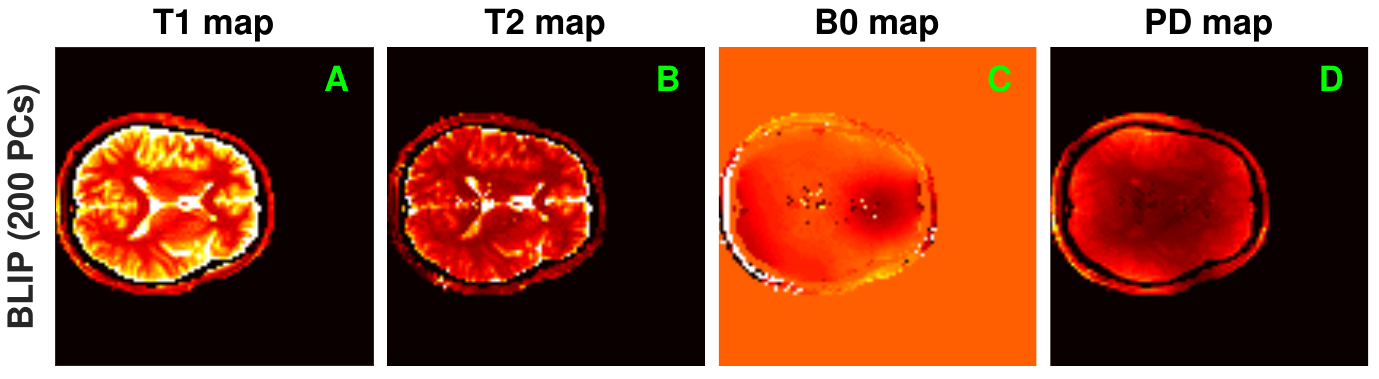} }
		\\
		\subfloat
		{\includegraphics[width=.95\textwidth]{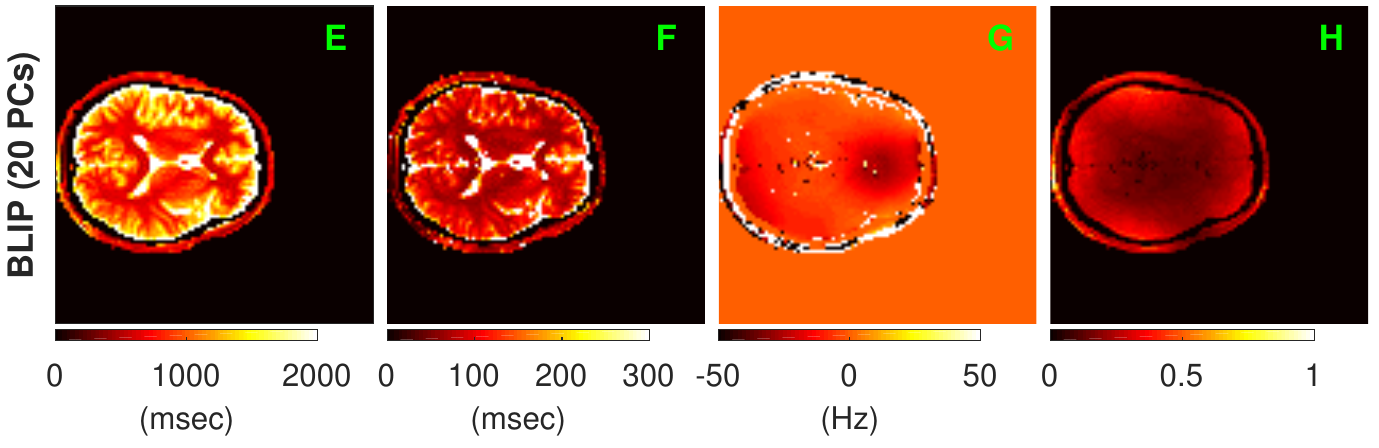} }	
		\caption{Reconstructed maps using BLIP algorithm (6-coil data). Significant subspace compression ($s=20$) causes overestimation of the $T2$ values (F) and distortion in the $B0$ map (G) e.g. around fat/muscle regions. A moderate compression ($s=200$) results in a reconstruction quality (A-D) comparable to that of using no temporal compression i.e. Figure~\ref{fig:coilcompare}(I-L).} \label{fig:svdcompare}
	\end{minipage}
\end{figure*}


\subsubsection{Missing high-resolution information and high-frequency artefacts}
As can be observed in Figure~\ref{fig:highres}(b), using iterative methods for spiral readouts may cause high-frequency artefacts in the estimated maps.  
 We would like to emphasize that this issue does not arise because of the deficiency of iterations. Indeed, the monotone decay of the measurement fidelity error implies that iterations improve data consistency as compared to the non-iterative TM scheme (Figure~\ref{fig:highres}(c)). As also highlighted in~\cite{AIRMRF}, after an initial rapid decay in the fidelity error a long epoch of slowly-decaying iterations will follow to recover high-resolution image features. 
 However, since spiral readouts do not sufficiently sample high-frequency k-space locations, solving~\eqref{eq:CS} may admit undesirable solutions with high-frequency artefacts which appear in the second epoch of iterations until convergence. These artefacts can be removed by either using a spatial-smoothing regularization\footnote{In \cite{AIRMRF} a low-pass filtering is used at each iteration to remove high-frequency artefacts. However, in conjunction with a non-convex matching step, such a sequential projection approach (i.e. for multiple constraints) would not guarantee the convergence of iterations.} or by reconstructing images in a lower spatial resolution. Here we take the latter approach and reconstruct volunteer images in $n=100\times100$ resolution for the rest of our experiments \textemdash instead of the $128\times128$ resolution maps shown in~\cite[Figure 3]{MRF} using the non-iterative TM. We also observe that with this update we require less iterations to converge.
 
\begin{table*}[t!]
	\centering
	\scalebox{.87}{
		\begin{tabular}{lc|ccc|ccc}
			\toprule[0.2em]
			Algorithm  & \multicolumn{1}{c}{ BLIP } &  \multicolumn{3}{c}{ CoverBLIP } &  \multicolumn{3}{c}{ KDBLIP }  \\
			\midrule[0.05em]
			ANNS parameter   & - & 0.1 & 0.2 & 0.4 & 256& 512 & 4096   \\
			\midrule[0.1em]
			& \multicolumn{7}{c}{Temporal compression $s=20$ }   \\
			\midrule[0.05em]
			Total search cost& 5.83e11 & 1.31e9 & 5.96e8 & 3.16e8 &2.74e9&5.38e9 & 3.58e10 \\
			Reconstruction time (sec)& 133.0 & 28.3 & 23.8 & 22.4 &25.3&26.5 & 48.9 \\
			Normalized fidelity error& 1.248e-1 & 1.266e-1 & 1.266e-1 & 1.267e-1 &1.275e-1&  1.273e-1 & 1.269e-1 \\
			\midrule[.1em]
			&\multicolumn{7}{c}{Temporal compression $s=200$ }   \\
			\midrule[0.05em]
			Total search cost& 6.55e12 & 6.97e10 & 2.28e10 & 9.91e9 &  4.88e10 & 9.68e10&6.76e11 \\
			Reconstruction time (sec)& 411.5 & 192.3 & 147.2 & 140.6 &  153.1 &156.5& 171.8 \\
			Normalized fidelity error& 1.285e-1 & 1.305e-1 & 1.306e-1 & 1.309e-1 &1.337e-1&  1.333e-1 & 1.328e-1\\
			\midrule[.1em]
			&\multicolumn{7}{c}{No temporal compression}   \\
			\midrule[0.05em]
			Total search cost& 2.91e13 & 3.84e11 & 1.37e11 & 6.76e10 &5.60e11& 1.33e12 & 8.11e12\\
			Reconstruction time (sec)& 1660.1 & 1014.7 & 652.4 & 684.1 &680.0&  825.6 & 846.6 \\
			Normalized fidelity error& 1.287e-1 & 1.307e-1 & 1.311e-1 & 1.320e-1 &1.490e-1&  1.458e-1& 1.414e-1 \\							
			\bottomrule[0.2em]
		\end{tabular}}
		\caption{Comparison between iterative methods  with/without using temporal subspace compression in terms of search cost, total runtime and normalized fidelity error.}\label{tab:comp}
	\end{table*}


\begin{figure*}
	\centering
	\begin{minipage}{\textwidth}
		\centering
		\subfloat
		{\includegraphics[width=.95\textwidth]{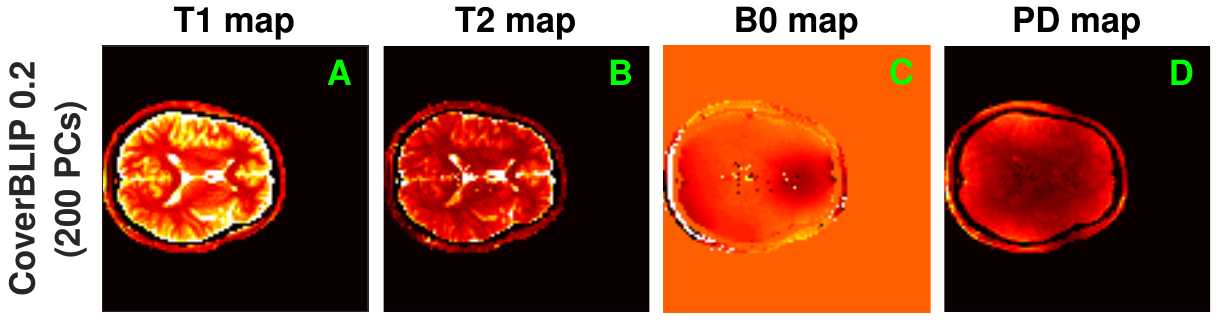} }
		\\
		\subfloat
		{\includegraphics[width=.95\textwidth]{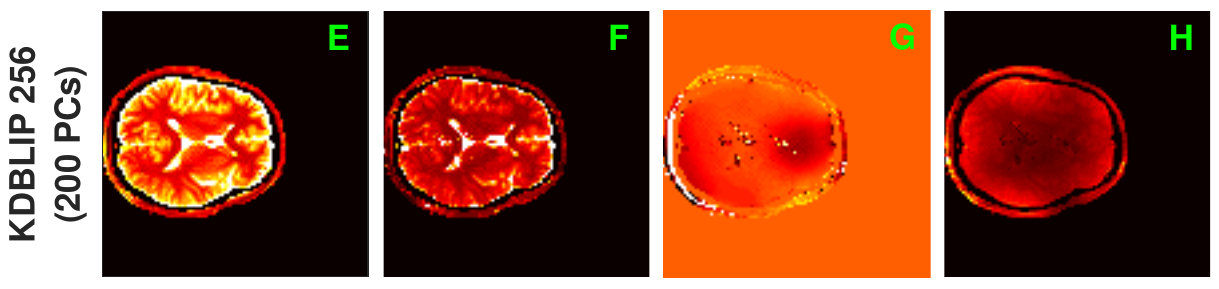} }
		\\
		\subfloat
		{\includegraphics[width=.95\textwidth]{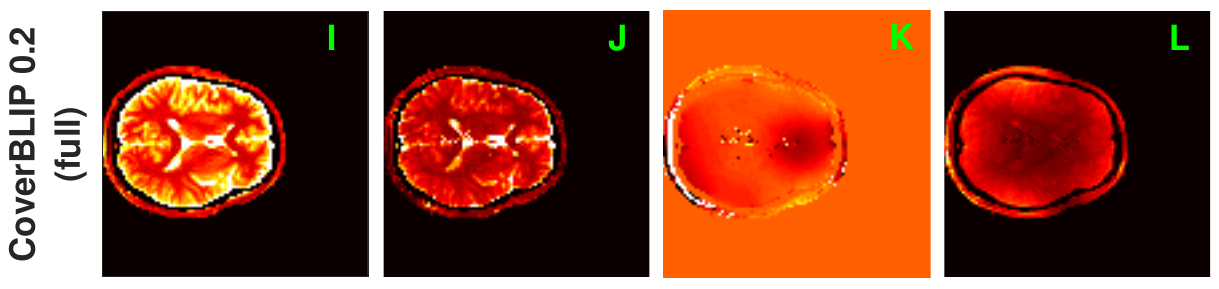} }	
		\\
		\subfloat
		{\includegraphics[width=.95\textwidth]{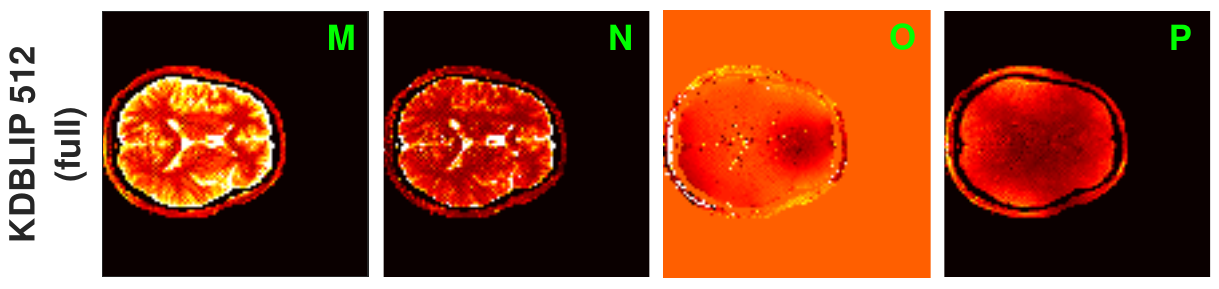} }	
		\\
		\subfloat
		{\includegraphics[width=.95\textwidth]{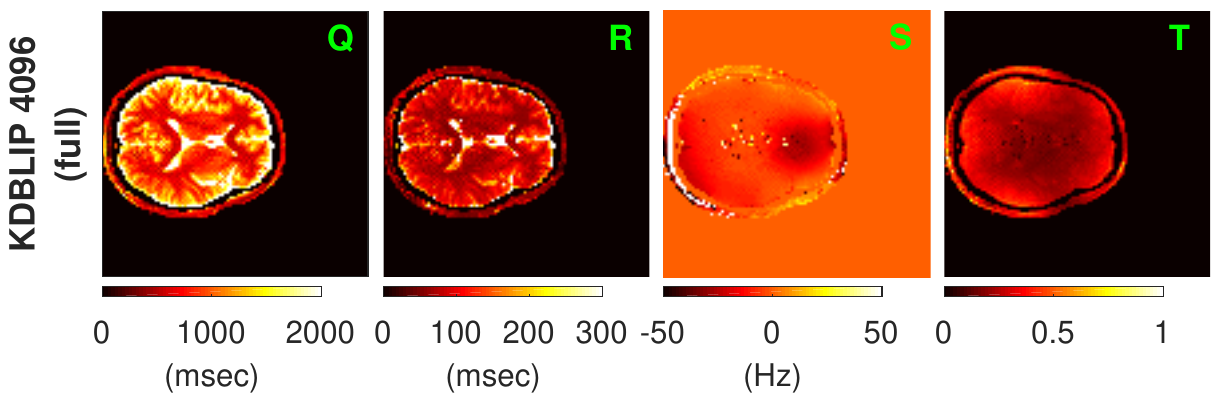} }		
		\caption{Reconstructed maps using inexact iterations of CoverBLIP and KDBLIP algorithms (6-coil data), tested for different search-dimension regimes i.e. with/without using temporal subspace compression.\label{fig:coverkdcompare}}
	\end{minipage}
\end{figure*}
 
\subsubsection{Results} The non-iterative TM algorithm performs reasonably well when all 32 coil/channel data are used (Figure~\ref{fig:coilcompare}(A-D)), 
supporting the fact that in data-rich regimes we may not need sophisticated inference algorithms~\cite{Chandra-Jordan}. 
To better highlight the advantage of iterations we select measurements from 6 coils that maximally cover the k-space. As can be observed in Figure~\ref{fig:coilcompare}(H,L), the recovered PD maps from the 6-coil data demonstrate weaker signal intensity in central and certain border regions as compared to the one obtained from the 32-coil data in Figure~\ref{fig:coilcompare}(D). Comparing Figure~\ref{fig:coilcompare}(B,C) to Figure~\ref{fig:coilcompare}(F,G) shows that TM reconstruction for the reduced 6-coil data introduces artefacts in both $T2$ and $B0$ maps around the Cerebrospinal Fluid (CSF) regions where the signal is weak. The iterative BLIP algorithm however corrects for this issue and works stably in low-data regime. The rest of our experiments focuses on the 6-coil k-space data. 

We next compare the performance of exact BLIP iterations using the subspace compression option. As we can see in Figure~\ref{fig:svdcompare}(F,G)  a significant SVD compression ($s=20$) causes overestimated $T2$ values and distorted $B0$ map (e.g. see fat/muscle regions) whereas, using a moderate compression ($s=200$) results in reconstruction quality comparable to that of using no subspace compression (compare the reconstructed maps in Figure~\ref{fig:svdcompare}(A-D) to  Figure~\ref{fig:coilcompare}(I-L)).
The reason is that the subspace of $s=200$ principal components can well represent the IR-BSSFP dictionary used in this experiment (see also related discussions in~\cite{SVDMRF}).

Table~\ref{tab:comp} compares the reconstruction performance of iterative methods BLIP, KDBLIP and CoverBLIP for different search dimension regimes with/without using subspace compression. 
The corresponding reconstructed maps can  be  also visually compared in Figure~\ref{fig:coverkdcompare}. The BLIP algorithm using exact brute-force searches achieves the lowest fidelity error but it requires the longest reconstruction time and highest search complexity. CoverBLIP with $\epsilon=0.2$ reports the best reconstruction time-accuracy (also total search cost) among all tested methods. CoverBLIP saves between 2 to 3 orders of magnitude in total search cost of BLIP with a comparable reconstruction accuracy (see the corresponding normalized fidelity errors in Table~\ref{tab:comp} and the recovered maps in Figure~\ref{fig:coverkdcompare}(A-D, I-L)). Importantly and unlike KDBLIP, this computational advantage is consistent for all three search-dimension regimes. We can observe in Figure~\ref{fig:coverkdcompare}(E-H) that by using subspace compression $s=200$, the KDBLIP algorithm with 256 checks outputs comparable parameter maps to that of CoverBLIP, however with 2-4 times more search cost. Runtimes reported for both methods in this case (Table~\ref{tab:comp}) are however similar because as previously pointed out we do not claim an optimal implementation of the cover tree searches used here. KDBLIP uses non-scalable tree searches and therefore without a dimensionality reduction \textemdash  even with a large number of checks = 4096, a longer runtime and 80x higher search cost than CoverBLIP\textemdash this algorithm fails to output artefact-free parameter maps (Figure~\ref{fig:coverkdcompare}(M-T)). More artefacts occur using smaller checks e.g. 512 or 256. In this experiment a moderate subspace compression turns out to be advantageous for all tested algorithms, but then it is a crucial step for using KD-tree searches. We empirically observed that KDBLIP starts reporting poor reconstruction time-accuracies when more than $350$ principal components are used. 

Comparing the overall runtimes in Table~\ref{tab:comp}, we note that CoverBLIP ($\epsilon=0.2$) achieves 2.5-6x fold acceleration compared to the BLIP algorithm which is less than what was reported for our previous synthetic data experiment in Section~\ref{sec:brainwebexpe}. The reason is that here we use multi-coil data and non-Cartesian k-space sampling where both  make the gradient updates become a non-trivial computational overhead for the iterations. Note that reconstructions from a non-Cartesian acquisition protocol requires computing slow NUFFT operations in each iteration. As a result, despite a significant reduction in the total search cost (i.e. projection steps) this advantage will be less pronounced in the overall runtime of CoverBLIP. We believe addressing this issue i.e. breaking down the cost of heavy gradient updates,  merits an independent line of future investigation beyond the scope of this work.

%% file: sections/conclusion.tex
\section{Conclusions and future directions}
\label{sec:conclusions}
We considered accelerating the iterative scheme for model-based MRF reconstruction and for this purpose we approximated the matched-filtering step in each iteration using cover tree's $(1+\epsilon)$-ANNS search scheme. For low-dimensional manifold  datasets cover trees offer appealing construction times, memory requirements and remarkably low  search complexities scaling logarithmic in terms of data population. With this motivation, we proposed the CoverBLIP algorithm which adopts such tree structures for fast iterative searches over large-size MRF dictionaries i.e. discrete manifold of Bloch responses parametrized by few NMR characteristics. Provided with a notion of (model-restricted) embedding 
we showed that the inexact iterations of CoverBLIP linearly  convergence toward a solution with the same order of accuracy as when using BLIP with exact brute-force searches. We also introduced an adaptive step-size scheme that guarantees local monotone convergence of CoverBLIP in the absence of  bi-Lipschitz embedding. We evaluated the performance of our proposed method on both synthetic and real-world MRF datasets using different sampling strategies, and we demonstrated that CoverBLIP is capable of achieving orders of magnitude acceleration in conducting the projection steps as compared to the exact iterations of BLIP. We also showed that CoverBLIP is a scalable algorithm able to maintain the gain in its time-accuracy performance in high-dimensional search spaces.

Future works include application of CoverBLIP to the emerging multi-parametric MRF problems with more complex dynamic responses encoding a larger number of NMR characteristics. In such cases and due to the inherent non-linearity of Bloch responses a low-dimensional subspace model of the dictionary would be prohibitively inaccurate,  
and one would rather need to resort on fast search schemes such as cover trees that are robust against the curse-of-dimensionality. Our current search implementation does not benefit from the considerable amount of inter-voxel correlations present in a \emph{query batch}. As shown in e.g.~\cite{curtinthesis} faster searches are possible 
by additionally building a dual (cover) tree on  the query batches. An interesting line of future work would adopt this idea to further accelerate CoverBLIP, however with more restricted choice of dual trees or batch sizes whose construction times would not bring a computational overhead throughout multiple iterations. 
Further extension to the present work could also focus on reducing the computational cost of the gradient updates for   non-Cartesian and multi-coil acquisition schemes. In this regard, a possible line of investigation would be the  application of \emph{randomized} iterative projected gradient algorithms (see e.g.~\cite{bottouSGD,svrg13,GPIS,restkatyusha}), where  iterations adopt cheap, unbiased and variance-reduced stochastic approximations of the true but computationally-intensive gradient updates.  

%

%% file: sections/proofs.tex
\section{Proof of the convergence results in Section~\ref{sec:proofs}}
\label{sec:proofs_supp}
In this part and in Section~\ref{sec:proof0} we provide the proof of Proposition~\ref{prop:stepbound} on the feasibility of using the adaptive shrinkage scheme to find a good step-size in a finite number of sub-iterations.  Section~~\ref{seq:proof1} includes the proof of  Theorem~\ref{th:inexactLS2} which establishes a near global convergence result i.e. a reconstruction guarantee, for the inexact IPG algorithm by using the adaptive step-size shrinkage scheme. In Section~\ref{sec:proof2} we provide the proof of Theorem~\ref{th:inexactmonotone} which guarantees monotone convergence of the inexact IPG in the absence of any embedding assumption.

\subsection{Proof of Proposition~\ref{prop:stepbound}}
\label{sec:proof0}
	Following the approximate projection definition~\eqref{eq:eproj}, each iteration of the inexact IPG algorithm~\eqref{eq:inIP2} produces $X^{k+1},X^{k}\in \Cc$. Therefore, according to the bi-Lipschitz property (Definition~\ref{def:bilip}) we have the following bounds on the values used in the step-size criteria~\eqref{eq:steprule} $\forall k$:
	\eq{\MM^{-1}\leq \frac{\norm{X^{k+1}-X^k}^2}{\norm{ \Aa(X^{k+1}-X^k) }^2 }\leq \mmx^{-1}.
	}
	As a result starting from any finite large step-size $\mu$ and after a finite number of divisions by a factor $\zeta>1$, $\mu_k$ reaches the lower bound and satisfies criteria~\eqref{eq:steprule}. The smallest possible step-size before stopping the shrinkage thus ranges in the interval $\left( (\zeta\MM)^{-1},\MM^{-1} \right]$ which implies the lower bound in~\eqref{eq:stepbound}. The largest possible $\mu_k$ is upper bounded by~\eqref{eq:steprule} which is always less than $\mmx^{-1}$.

\subsection{Proof of Theorem~\ref{th:inexactLS2}}
\label{seq:proof1}
We start from a similar argument as for the proof of~\cite[Theorem~2]{inexactipg-tit} however we do not assume $\mu_k\leq 1/\MM$ as there or in Theorem~\ref{th:inexactLS1} of this paper. 
By setting $g :=2\Aa^H(\Aa(X^{k})-Y)$ it follows that
\begin{align*} 
\norm{Y-\Aa(X^{k+1})}^2-\norm{Y-\Aa(X^{k})}^2	&= \langle X^{k+1}-X^{k},g \rangle +\norm{\Aa(X^{k+1}-X^{k})}^2 \\
&\leq \langle X^{k+1}-X^{k},g \rangle + \frac{1}{\mu_k} \norm{X^{k+1}-X^{k}}^2\\
& = \frac{1}{\mu_k} \norm{X^{k+1}-X^{k}+\frac{\mu_k}{2} g }^2 - \frac{\mu_k}{4} \norm{g}^2,
\end{align*}
where the inequality follows from the step size rule~\eqref{eq:steprule}. Due to the update rule of algorithm~\eqref{eq:inIP2} with the inexact projection defined in \eqref{eq:eproj}, we have
\begin{align*}
\norm{X^{k+1}-X^{k}+\frac{\mu_k}{2} g }^2 &=  \norm{\pp_{\Cc}(X^{k}-\frac{\mu_k}{2} g)-X^{k}+\frac{\mu_k}{2} g }^2 \\
&\leq  (1+\epsilon)^2\norm{\Pp_{\Cc}(X^{k}-\frac{\mu_k}{2} g)-X^{k}+\frac{\mu_k}{2} g }^2 \\
&\leq \norm{\xgt-X^{k}+\frac{\mu}{2} g }^2 +\phi(\epsilon)^2 \frac{\mu_k^2}{4}\norm{g}^2
\end{align*}
where $\phi(\epsilon):=\sqrt{2\epsilon+\epsilon^2}$. For the last inequality we replace $\Pp_{\Cc}(X^{k}-\frac{\mu}{2} g)$ with two feasible points $\xgt,X^{k}\in \Cc$. 
Therefore we can write
\begin{align}
\norm{Y-\Aa(X^{k+1})}^2-\norm{Y-\Aa(X^{k})}^2 \nonumber	
&\leq \frac{1}{\mu_k} \norm{\xgt-X^{k}+\frac{\mu_k}{2} g }^2 - \frac{\mu_k}{4} \norm{g}^2 + {\phi(\epsilon)^2}\frac{\mu_k}{4}\norm{g}^2 \nonumber \\
&= \langle \xgt-X^{k},g \rangle + \frac{1}{\mu_k} \norm{\xgt-X^{k}}^2 
+{\phi(\epsilon)^2}\frac{\mu_k}{4}\norm{g}^2.
 \label{eq:p1b2}
\end{align}
According to the bi-Lipschitz property (Definition~\ref{def:bilip}) we have the following bound:
\begin{align*} 
\langle \xgt-X^{k},g \rangle 	&= w^2-\norm{Y-\Aa(X^{k})}^2 -\norm{\Aa(\xgt-X^{k})}^2 \\
&\leq w^2 -\norm{Y-\Aa(X^{k})}^2 -\mmx\norm{\xgt-X^{k}}^2,
\end{align*}
where $w=\norm{ Y-\Aa(\xgt)}$. Replacing this bound in \eqref{eq:p1b2} yields
\begin{align}
\norm{Y-\Aa(X^{k+1})}^2
\leq \left(\frac{1}{\mu_k}-\mmx \right)\norm{X^{k}-\xgt}^2 
+ {\phi(\epsilon)^2}\frac{\mu_k}{4}\norm{g}^2
+w^2. \label{eq:p1b3}
\end{align}
On the other hand the following lower bound holds:
\begin{align*}
\norm{Y-\Aa(X^{k+1})}^2
&= \norm{\Aa(X^{k+1}-\xgt)}^2+w^2-2\langle y-\Aa\xgt, \Aa(X^{k+1}-\xgt)\rangle\\
&\geq \norm{\Aa(X^{k+1}-\xgt)}^2+w^2-2w \norm{\Aa(X^{k+1}-\xgt)}\\
& \geq \mmx\norm{X^{k+1}-\xgt}^2+w^2-2w \sqrt{\MM}\norm{X^{k+1}-\xgt}\\
&= \left(\sqrt{\mmx}\norm{X^{k+1}-\xgt}- \sqrt{\frac{\MM}{\mmx}}w\right)^2 -(\frac{\MM}{\mmx}-1)w^2.
\end{align*}
The first and second inequalities use Cauchy-Schwartz and the bi-Lipschitz property, respectively. Using this bound in~\eqref{eq:p1b3} yields
\begin{align*}
\left(\sqrt{\mmx}\norm{X^{k+1}-\xgt}- \sqrt{\frac{\MM}{\mmx}}w\right)^2
&\leq \left(\frac{1}{\mu_k}-\mmx \right)\norm{X^{k}-\xgt}^2 + \phi(\epsilon)^2 \frac{\mu_k}{4}\norm{g}^2+\frac{\MM}{\mmx}w^2 \\
&\leq \left(\sqrt{\frac{1}{\mu_k}-\mmx} \norm{X^{k}-\xgt} + 
\phi(\epsilon) \frac{\sqrt{\mu_k}}{2}\norm{g}
+\sqrt{\frac{\MM}{\mmx}}w \right)^2.
\end{align*}
The last inequality assumes $\forall k, \mu_k\leq \mmx^{-1}$ which holds due to the upper bound~\eqref{eq:stepbound} on chosen step size in Proposition~\ref{prop:stepbound}. 
On the other hand by triangle inequality we have
\begin{align*}
\norm{g}&\leq 2\norm{\Aa^H(\Aa(X^{k}-\xgt))}+2\norm{\Aa^H(Y-\Aa(\xgt))}\\
&\leq 2\sqrt{ 1/\mu_k}\vertiii{\Aa}\norm{(X^{k}-\xgt)}+2\vertiii{\Aa}w,
\end{align*}
where the last inequality follows from the step size criteria~\eqref{eq:steprule}.
As a result we deduce that
\begin{align*}
\norm{X^{k+1}-\xgt}&\leq
\left(\sqrt{\frac{1}{\mu_k\mmx}-1}+ \phi(\epsilon)\frac{\vertiii{\Aa}}{\sqrt{\mmx}}\right) \norm{X^{k}-\xgt} 
+ \left( 2\frac{\sqrt{\MM}}{\mmx}+ \phi(\epsilon)\sqrt{\frac{\mu_k}{\mmx}} \vertiii{A} \right)w\\
&\leq
\left(\sqrt{\frac{\zeta\MM}{\mmx}-1}+ \phi(\epsilon)\frac{\vertiii{\Aa}}{\sqrt{\mmx}}\right) \norm{X^{k}-\xgt} 
+ \left( 2\frac{\sqrt{\MM}}{\mmx}+ \phi(\epsilon)\frac{\vertiii{A}}{\mmx}  \right)w\\
&\leq
\left(\sqrt{\frac{\zeta\MM}{\mmx}-1}+ \delta\right) \norm{X^{k}-\xgt} 
+ \left( 2\frac{\sqrt{\MM}}{\mmx}+ \frac{\delta}{\sqrt{\mmx}} \right)w.
\end{align*}
The second inequality uses both lower and upper bounds \eqref{eq:stepbound} on the adaptive step size, and the last inequality follows from the theorem's assumption 
$\phi(\epsilon)\vertiii{\Aa}/\sqrt{\mmx}\leq\delta$.  Applying this bound  recursively completes the proof:
\eq{
	\norm{X^{k}-\xgt}\leq  \rho^k \norm{\xgt} +  \frac{\kappa_w}{1-\rho}w
}
for $\kappa_w$ defined in Theorem \ref{th:inexactLS2}. The condition for convergence is $\rho<1$ which implies $\delta<1$ and the following conditioning between the bi-Lipschitz constants $\zeta \MM < \mmx+(1-\delta)^2\mmx$.

\subsection{Proof of Theorem~\ref{th:inexactmonotone}}
\label{sec:proof2}
	The proof is simple in the light of viewing the IPG algorithm as a majorization-minimization. Consider the following cost function at iteration $k$: 
	\eql{
		\Ll(X) := \mu_k\norm{Y-\Aa(X)}^2 + \norm{X-X^k}^2 - \mu_k\norm{\Aa(X-X^k)}^2.
	}
	On one hand due to the step-size criteria~\eqref{eq:steprule} we have $\mu_k\norm{Y-\Aa(X^{k+1})}< \Ll(X^{k+1})$ for any $X^k \neq X^{k+1}$. Also by definition it holds  $\Ll(X^k)=\mu_k\norm{Y-\Aa(X^{k})}$. On the other hand note that $\Ll(X) = \mu_k \norm{X-Z^k}^2+\text{const}$ for some constant $\text{const}$, and thus according to~\eqref{eq:nonexpansive} for any non-expansive projection updates $X^{k+1}=\pp_\Cc(Z^k)$ we have $\Ll(X^{k+1})\leq \Ll(X^k)$. 
	Chaining these inequalities yields
	\eql{
		\norm{Y-\Aa(X^{k+1})}<\norm{Y-\Aa(X^{k})} \qquad \forall k, X^k\neq X^{k+1}. }
	which (together with the fact that the cost function $\norm{Y-\Aa(X)}\geq 0$ is lower bounded) completes the proof.

%% file: sections/dcf.tex
\section{A note on using density compensation for preconditioning the MRF reconstruction with spiral readouts}
\label{sec:dcf}

\begin{figure*}
	\centering
	\begin{minipage}{\textwidth}
		\centering
		\subfloat
		{\includegraphics[width=.95\textwidth]{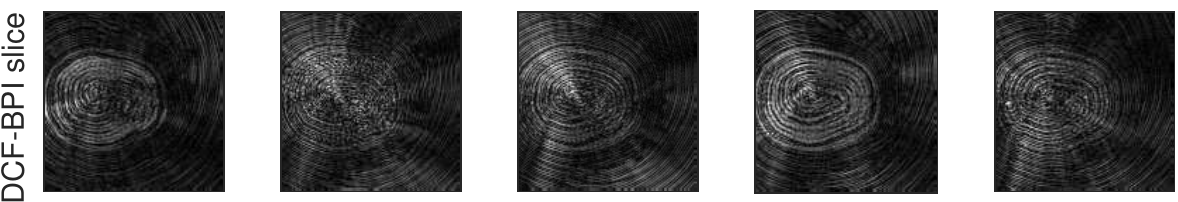} }
		\\
		\subfloat
		{\includegraphics[width=.95\textwidth]{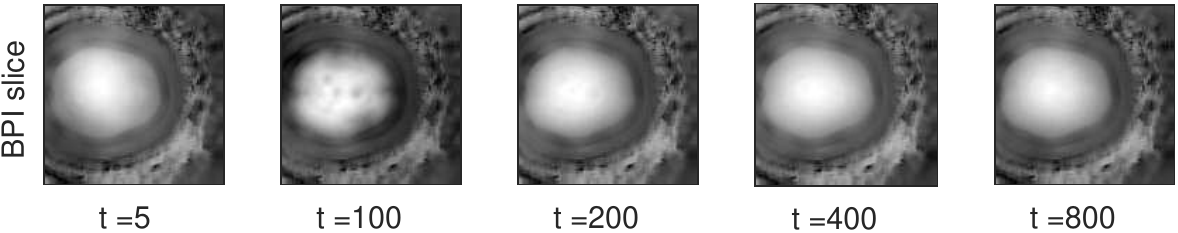} }
		\\
		(a) Back-projected images with (top row) and without (bottom row) DCF weighting 
		\\
		\subfloat
		{\includegraphics[width=.45\textwidth]{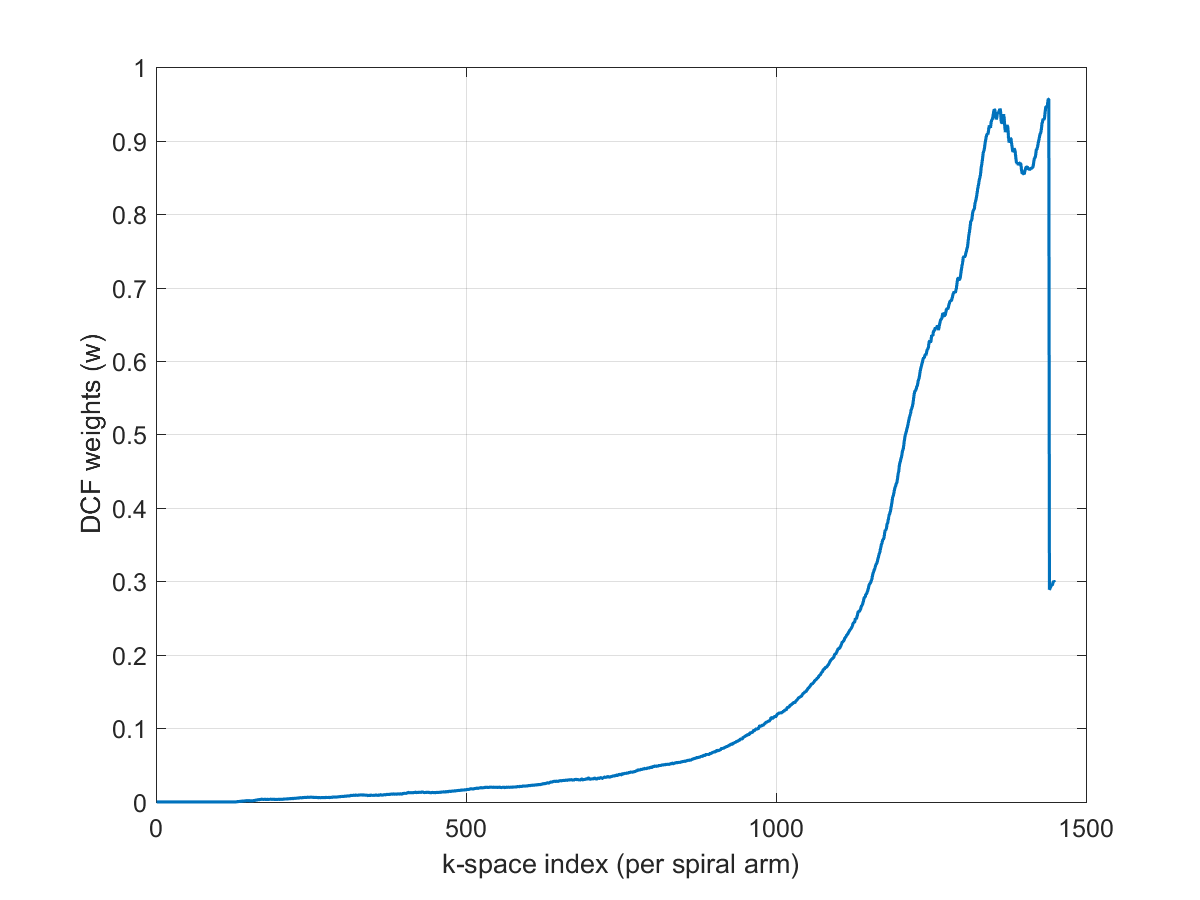} }
		\\
		(b) DCF profile across one spiral arm 	
		\caption{Figure (a) demonstrates five BPI temporal slices  reconstructed with and without DCF compensation i.e. $X' = \Aa^H\left(\text{diag}(w)Y\right)$ and $X' = \Aa^H(Y)$, respectively. Figure (b) shows the DCF weights computed for the spiral readouts used in this experiment. Indices concentrated near the centre of k-space have relatively smaller weights.\label{fig:BPI_scan}}
	\end{minipage}
\end{figure*}

Spiral readouts acquire much denser collection of samples from the centre of k-space than the outer regions. As a result the forward operator $\Aa$ becomes more ill-conditioned as compared to e.g. a Cartesian acquisition. In this case if we perform an iterative reconstruction such as \eqref{eq:blip} the progress in each iteration will be very small. As can be observed in Figure~\ref{fig:BPI_scan}(a), applying the adjoint operator on the k-space data $X' = \Aa^H(Y)$ results in heavily blurred images in the first iteration. Such slow convergence combined with costly NUFFT updates at each iteration makes the reconstruction extremely time-consuming. One fix to this issue \textemdash which we adopt in our experiments in Section~\ref{sec:expe_invivo} of the main article \textemdash is to use a \emph{weighted} least squares loss for the fidelity term with smaller weights for the central k-space locations. The objective of \eqref{eq:CS} updates to
\eq{
	\sum_{t=1}^L \norm{Y_t-P_{\Omega_t} FS(X_t)}_\Ww^2,
}
where the weighted norm is defined as $\norm{a}_\Ww^2 := \sum_i w_i|a_i|^2$, and the weights $w_i>0$ are derived from calculating the Density Compensation Function (DCF) for a given spiral trajectory~\cite{dcf-voronoi}. Figure~\ref{fig:BPI_scan}(b) illustrates DCF weights for the sampling trajectory used in this experiment. Following the update in the objective function and the corresponding gradient expression, the iterations of BLIP in~\eqref{eq:blip} take the following form:
\eql{\label{eq:blip_dcf}
	X^{k+1} = \Pp_\Cc \left(X^k - \mu^k \Aa^H\left(\text{diag}(w)\left(\Aa(X^k)-Y \right)\right)\right). 
}
Similar update applies to CoverBLIP \eqref{eq:inIP2}, KDBLIP and their SVD dimension-reduced variants when using spiral readouts. 
Figure~\ref{fig:BPI_scan}(a) illustrates the weighted back projected images (BPI) i.e. $X' = \Aa^H\left(\text{diag}(w)Y\right)$ in the first iteration, where we can observe the blurring has been removed  (the under-sampling artefacts however remain).  DCF reweighing significantly improves the conditioning of Hessian $\Aa^H\text{diag}(w)\Aa$ compared to that of $\Aa^H\Aa$ and results in much larger chosen step-size and faster progress during the iterations i.e. faster convergence. Similar to the Carteian sampling we initialize the step-size by the compression factor $\mu=n/m$ and we empirically observe that this choice satisfies the criteria~\eqref{eq:steprule} for most of the iterations and for the rest one or two shrinkage sub-iterations suffices.